\pdfoutput=1

\documentclass[11pt]{article}

\usepackage[]{ACL2023}

\usepackage{times}
\usepackage{latexsym}
\usepackage{multirow}
\usepackage{booktabs}
\usepackage{makecell}
\usepackage{amssymb}
\usepackage{amsmath}
\usepackage{pifont}
\usepackage{graphicx}
\usepackage{tabularx}
\usepackage{bbm}
\usepackage{bm}
\usepackage{arydshln}
\usepackage{xcolor}
\usepackage{xspace}

\graphicspath{{./figures/}}

\usepackage[T1]{fontenc}

\usepackage[utf8]{inputenc}

\usepackage{microtype}

\usepackage{inconsolata}

\title{Where's the Point? Self-Supervised Multilingual Punctuation-Agnostic\\ Sentence Segmentation}

\newcommand{\ourmethod}{WtP\xspace}
\newcommand{\ourunsup}{WtP${_\textsc{u}}$\xspace}
\newcommand{\ourthresholdtuned}{WtP${_\textsc{t}}$\xspace}
\newcommand{\ourpuncttuned}{WtP${_\textsc{punct}}$\xspace}
\newcommand{\spacysent}{SpaCy${_\textsc{sent}}$\xspace}
\newcommand{\spacydp}{SpaCy${_\textsc{dp}}$\xspace}
\newcommand{\ersatzunsup}{Ersatz${_\textsc{u}}$\xspace}
\newcommand{\canines}{$\textsc{Canine-S}$\xspace}
\renewcommand{\textapprox}{\raisebox{0.5ex}{\texttildelow}}
\definecolor{ourred}{HTML}{E13342}
\definecolor{ourblue}{HTML}{6495ed}

\newcommand{\subrparagraph}[1]{\vspace{1.2mm}\noindent\textbf{#1}}
\newcommand{\rparagraph}[1]{\vspace{1.8mm}\noindent\textbf{#1.}}
\newcommand{\sparagraph}[1]{\vspace{0.0mm}\noindent\textbf{#1.}}

\usepackage{todonotes}

\makeatletter
\newcommand*\iftodonotes{\if@todonotes@disabled\expandafter\@secondoftwo\else\expandafter\@firstoftwo\fi}
\makeatother

\author{Benjamin Minixhofer\thanks{\, Work done during the time BM interned at Cohere.}$ \, \, ^1$ \, \, Jonas Pfeiffer\thanks{\, Equal senior authorship.}$\, \, ^2$  \, \, Ivan Vuli\'c$^{\dagger 3}$ \\
$~~~~~~~^1$Cohere for AI~~~~~~~$^2$Google DeepMind~~~~~~~$^3$University of Cambridge 
}

\begin{document}
\maketitle
\begin{abstract}

Many NLP pipelines split text into sentences as one of the crucial preprocessing steps. Prior sentence segmentation tools either rely on punctuation or require a considerable amount of sentence-segmented training data: both central assumptions might fail when porting sentence segmenters to diverse languages on a massive scale. In this work, we thus introduce a \textit{multilingual} \textit{punctuation-agnostic} sentence segmentation method, currently covering 85 languages, trained in a self-supervised fashion on unsegmented text, by making use of newline characters which implicitly perform segmentation into paragraphs. 
We further propose an approach that adapts our method to the segmentation in a given corpus by using only a small number (64-256) of sentence-segmented examples. 
The main results indicate that our method outperforms all the prior best sentence-segmentation tools by an average of 6.1\% F1 points. Furthermore, we demonstrate that proper sentence segmentation has a point: the use of a (powerful) sentence segmenter makes a considerable difference for a downstream application such as machine translation (MT). By using our method to match sentence segmentation to the segmentation used during training of MT models, we achieve an average improvement of  2.3 BLEU points over the best prior segmentation tool, as well as massive gains over a trivial segmenter that splits  text into equally sized blocks.

\end{abstract}

\section{Introduction}
\label{sec:introduction}
Sentences are ubiquitous in NLP. Many datasets are made up of annotated sentences~\citep[][\textit{inter alia}]{de-marneffe-etal-2021-universal, aharoni-etal-2019-massively, conneau-etal-2018-xnli} and models often expect individual sentences as input \citep[][\textit{inter alia}]{reimers-gurevych-2019-sentence, reimers-gurevych-2020-making, liu-etal-2021-fast, tiedemann-thottingal-2020-opus}. This mandates a need for tools to segment text into sentences: a requirement that typically slips under the radar of many modern NLP systems \cite{wicks-post:2022:WMT}.

\begin{table}[t!]
\small
\def\arraystretch{1.0}
\centering
\setlength\tabcolsep{4pt}
\begin{tabularx}{\linewidth}{lrX}
\toprule
\thead{Collection} & \thead{Sentences}\\
\cmidrule(lr){2-2}
UD & \scalebox{0.9}{\makecell[Xt]{This is the high season for tourism;\textcolor{red}{\textbf{ |}} between December and April few people visit and many tour companies and restaurants close down.}}\\
\cmidrule(lr){2-2}
OPUS100 & \scalebox{0.9}{\makecell[Xt]{'I couldn't help it,' said Five, in a sulky tone; 'Seven jogged my elbow.'\textcolor{red}{\textbf{ | }}On which Seven looked up and said, 'That's right, Five! Always lay the blame (...)!'}}\\
\cmidrule(lr){2-2}
Ersatz & \scalebox{0.9}{\makecell[Xt]{"A lot of people would like to go back to 1970," before program trading, he said.\textcolor{red}{\textbf{ |}} "I would like to go back to 1970.\textcolor{red}{\textbf{ |}} But we're not going back (...)"}}\\
\bottomrule
\end{tabularx}
\caption{Example sentences of different collections. The pipe ('\textcolor{red}{\textbf{|}}') indicates sentence boundaries.}
\label{table:examples}
\end{table}
\begin{figure}[t!]
\centering
\hspace{-0.15cm}
\includegraphics[width=0.99\linewidth+0.3cm]{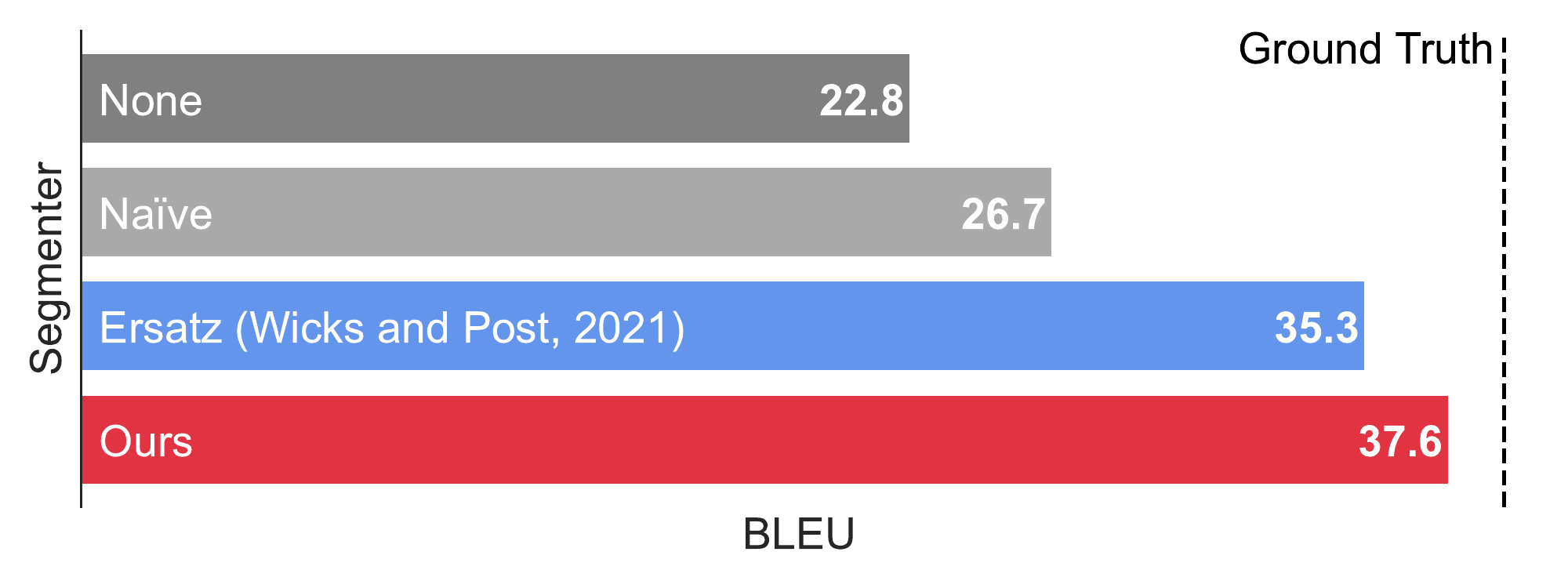}
\caption{Impact of sentence segmentation on BLEU scores in MT. Full details in Table~\ref{table:extrinsic}.}
\label{figure:extrinsic_brief}
\end{figure}

Theoretically, a sentence can be defined as a sequence of grammatically linked words conveying a complete thought~\citep{sweet2014new}.
In practice, there is ambiguity in what can be considered one sentence, as illustrated in Table~\ref{table:examples}. Do nested syntactic structures (e.g. through quotation marks) make up \textit{one} sentence, or \textit{multiple} ones? What about parentheses and enumerations? Sometimes, even colons and semicolons are considered sentence boundaries. In addition to the ambiguity in what makes up a sentence, there is practical difficulty in devising a tool to segment text into sentences. In many languages, punctuation is not limited to appearing at sentence boundaries, being used also for, e.g., acronyms and abbreviations. Other languages, such as Thai, do not use punctuation at all. In languages which do use punctuation, noisy user-generated text may still lack consistent punctuation \citep{kagan1980run}.

As surveyed later in \S\ref{sec:rw}, tools for sentence segmentation typically rely on sentence boundaries to occur exclusively at punctuation marks. This makes them applicable only to well-punctuated text in languages with sentence-ending punctuation. Some existing sentence segmentation tools do not rely on punctuation \citep{zhou-etal-2016-word, honnibal-johnson-2015-improved}; they, however, need sentence-segmented training data, which makes them difficult to apply to low-resource setups and languages. 

In order to address these core challenges, in this work we present a fully self-supervised sentence segmentation method which does not rely on punctuation, and is thus applicable to a wide spectrum of languages and different corpora. We pragmatically define a sentence as a sequence of characters which could plausibly be followed by a newline, following the intuition that paragraph breaks can never occur within a sentence. We then train a bidirectional character-level language model (ChLM) on text stripped of newline characters to predict, for each character, whether it was followed by a newline in the original text. A single configurable
threshold then determines whether each newline-likelihood score should or should not be treated as a sentence boundary. Our method, though self-supervised, on average matches the performance of the best prior (supervised) segmentation tools.

In addition, we take into account the fact that sentence segmentation is subjective and might be considered corpus-specific. To this end, we devise an auxiliary punctuation-prediction objective which allows adapting our model to the sentence segmentation in a given corpus in a data-efficient way. This leads to an improvement of an average 6.1\% F1 points over prior tools. We find that, while the precise definition of a sentence may not be important, consistency between training and inference is crucial: using our method to match sentence segmentation to the segmentation used during training leads to an improvement of an average 2.3 points in BLEU score of machine translation systems across 14 languages, as summarised in Figure~\ref{figure:extrinsic_brief}. %

\rparagraph{Contributions}
\textbf{1)} We introduce \textit{'Where's the Point'} (\ourmethod{}),
a method for self-supervised sentence segmentation without relying on punctuation and without language-specific assumptions. \textbf{2)} We present a data-efficient way to adapt \ourmethod{} models to the sentence segmentation in a given corpus using a small number (e.g., 64) of sentence-segmented examples. \textbf{3)} 
We train state-of-the-art \ourmethod{} models in five different sizes covering 85 languages. Our code and models are publicly available at \href{https://github.com/bminixhofer/wtpsplit}{\nolinkurl{github.com/bminixhofer/wtpsplit}}.

\section{Related Work and Baseline Systems}
\label{sec:rw}
\begin{table}[t!]
\small
\def\arraystretch{1.0}
\centering
\setlength\tabcolsep{3pt}
\begin{tabular}{llcr}
\toprule
\thead{\bf Type} & \thead{\bf Method} & \thead{\bf NP?} & \thead{\bf Reference}\\
\cmidrule(lr){1-4}
\multirow{3}{*}{\em 1. RB} & Moses & &\citet{koehn-etal-2007-moses}\\
& \spacysent & &\citet{Honnibal_spaCy_Industrial-strength_Natural_2020}\\
& PySBD & &\citet{sadvilkar-neumann-2020-pysbd}\\
\cmidrule(lr){2-4}
\multirow{5}{*}{\em 2. SS} & Riley & &\citet{riley-1989-applications}\\
& Satz & &\citet{palmer-hearst-1997-adaptive}\\
& Splitta & & \citet{gillick-2009-sentence}\\
& \spacydp  & \checkmark & \citet{honnibal-johnson-2015-improved}\\
& Ersatz & &\citet{wicks-post-2021-unified}\\
\cmidrule(lr){2-4}
\multirow{3}{*}{\em 3. US} & Punkt & &\citet{kiss-strunk-2006-unsupervised}\\
& \ersatzunsup & &\citet{wicks-post-2021-unified}\\
& \ourmethod & \checkmark & Ours\\
\bottomrule
\end{tabular}
\caption{Taxonomy of sentence segmentation methods into \textit{rule-based} (RB), \textit{supervised statistical} (SS) and \textit{unsupervised statistical} (US) approaches and whether they can segment text with no punctuation (NP?).}
\label{table:taxonomy}
\end{table}

\sparagraph{Sentence Segmentation} Work on sentence segmentation can be divided into \textbf{1)} rule-based, \textbf{2)} supervised statistical, and \textbf{3)} unsupervised statistical approaches. Table~\ref{table:taxonomy} shows a taxonomy of sentence segmentation methods, discussed in what follows.

\subrparagraph{1. Rule-Based Methods} rely on handcrafted rules to segment text into sentences. %
The sentence segmenters in Moses~\citep{koehn-etal-2007-moses} and SpaCy~\citep[\spacysent;][]{Honnibal_spaCy_Industrial-strength_Natural_2020} split on every punctuation character, unless it occurs within a handcrafted set of exceptions (e.g., abbreviations and acronyms). PySBD \citep{sadvilkar-neumann-2020-pysbd} uses a set of exceptions as well as regular expression rules. Rule-based approaches %
require extensive manual effort for every language, which makes scaling to many languages difficult.

\subrparagraph{2. Supervised Statistical Methods} use a corpus of already sentence-segmented text to learn segmentation. One of the first approaches in this area was by \citet{riley-1989-applications}, where they decide for each punctuation mark in a text, whether it constitutes a sentence boundary or not. They do so by learning a decision tree from lexical features of the context around the punctuation mark. Satz~\citep{palmer-hearst-1997-adaptive} and Splitta~\citep{gillick-2009-sentence} follow the same paradigm, but use a neural network with part-of-speech features, and an SVM with lexical features, respectively. The above approaches all suffer from the same limitation: the set of plausible sentence boundaries is made up of the set of punctuation marks, that is, a non-punctuation character can never make up a sentence boundary. This is a major restriction especially for text which is not well-punctuated, and text in languages which do not require punctuation (e.g. Thai). 

The dependency parser in the SpaCy library~\citep[\spacydp;][]{Honnibal_spaCy_Industrial-strength_Natural_2020} is among the first to lift this restriction. \spacydp jointly learns dependency parsing and sentence segmentation on a labelled corpus using a transition-based parser without special treatment of punctuation~\citep{honnibal-johnson-2015-improved}. Ersatz~\citep{wicks-post-2021-unified} modernizes \citet{riley-1989-applications}'s paradigm by using a Transformer~\citep{NIPS2017_3f5ee243} with subwords as context around punctuation. This again requires sentence boundaries to exclusively occur at punctuation marks. 

\subrparagraph{3. Unsupervised Statistical Methods} aim to learn segmentation using raw unsegmented text only. Punkt~\citep{kiss-strunk-2006-unsupervised} identifies abbreviations, initials and ordinal numbers in an unsupervised way by using character length and internal punctuation, among other features. All punctuation marks occurring outside of these are considered sentence boundaries. More recently, in addition to their supervised model, \citet{wicks-post-2021-unified} introduce a self-supervised model which follows the same paradigm as Ersatz, but is instead trained on punctuation preceding paragraph breaks.
This allows training without any labelled data, since newline characters naturally segment text into paragraphs. 
We refer to this model as \ersatzunsup{}.

Our method is most closely related to \ersatzunsup{}. We also use newlines (i.e. paragraph breaks) as a signal to learn segmentation. In contrast to \ersatzunsup{}, our method does not require punctuation. Also related, though specific to English, is \citet{moore2021indirectly} which uses n-gram occurences around paragraph breaks to predict sentence boundaries.

\rparagraph{Character-Level Pretrained LMs} 
Pretraining LMs on a large amount of text in a self-supervised way before training on the target task was a paradigm shift induced by BERT~\citep{devlin-etal-2019-bert}. Pretrained LMs typically use the Transformer architecture~\citep{NIPS2017_3f5ee243}. Pretrained LMs often represent the input as subword tokens~\citep{kudo-richardson-2018-sentencepiece, sennrich-etal-2016-neural}. However, recent efficiency improvements have enabled directly using characters as the input~\citep{clark-etal-2022-canine, tay2022charformer}. Character-level LMs (ChLMs) are well suited for multilingual sentence segmentation since (i) merging characters into subword tokens restricts sentence boundaries to end-of-token positions and (ii) subword-based tokenization leads to problems of vocabulary allocation in multilingual LMs \citep{rust-etal-2021-good}.

\section{Method}
\label{sec:method}
In light of the ambiguity in what makes up a sentence, we resort to the following definition:%
\begin{equation}
\parbox{7.1cm}{(D1) \em A sentence is any sequence of characters which could plausibly be followed by a newline.} \notag
\label{definition:sentence}
\end{equation}

This pragmatically driven definition corresponds closely to our intuitive understanding of a sentence due to two statements we assume to be true about sufficiently clean text: (i) a newline can generally not occur within a sentence and (ii) a newline can generally occur after any sentence\footnote{Together, this means that newlines are a \textit{high-precision, low-recall indicator of sentence boundaries} in the same way that hyphens are a high-precision, low-recall indicator of compound constituent boundaries \citep{minixhofer2023compoundpiece}.}.

This definition turns sentence segmentation into a character-level language modeling task. However, a causal language modeling objective as used in contemporary generative LMs \citep{NEURIPS2020_1457c0d6, radford2019language, radford2018improving} would restrict the model to unidirectional context. We devise a corruption method to allow using a bidirectional LM to model \textit{newline-likelihood}. Let $\bm{c}$ denote the sequence of characters making up some corpus.
$\bm{c}$ is preprocessed by stripping consecutive newline characters, and adding a space after every newline in languages which use whitespace to separate sentences. First, we corrupt the text by removing newline characters ($\mathtt{\backslash n}$) from $\bm{c}$, resulting in $\bm{x}$:
\begin{equation}
\bm{x} = \{c_i \:|\: c_i \in \bm{c}, c_i \neq \mathtt{\backslash n}\}.
\label{equation:corruption}
\end{equation}
The target is to identify which characters were followed by a newline in the original sequence:\footnote{Note that $c_i$ and $c_{i+1}$ index into the original sequence $\bm{c}$.}
\begin{equation}
\bm{y} = \left\{\!
\begin{array}{l}
\!1 \text{\; if $c_{i+1} = \mathtt{\backslash n}$}\\
\!0 \text{\; otherwise}
\end{array}
\:|\: c_i \in \bm{x}
\right\}
\end{equation}
Let the contextualized representations $\bm{h} = f_{\theta}(\bm{x})$ and predictions $\bm{\hat{y}} = \text{sigmoid}(g_\theta(\bm{h}))$ be produced by a character-level language model $f_{\theta}$ and a prediction head $g_\theta$ parameterized
by $\theta$. The loss is the standard binary cross-entropy between $\bm{y}$ and $\bm{\hat{y}}$:
\begin{equation}
\mathcal{L}_{\theta}^{\mathrm{main}} = -\frac{1}{|\bm{y}|} \sum_{i=0}^{|\bm{y}| - 1} y_i \log \hat{y}_i + (1 - y_i) \log(1 - \hat{y}_i) \notag
\end{equation}
The output $\bm{\hat{y}}$ can be interpreted as an estimate for the probability of a newline to occur after any character. This objective is comparable to objectives used in text-editing models~\citep{malmi-etal-2022-text}.\footnote{A model trained with this objective can be interpreted as a text-editing model with the singular operation \textit{'insert newline'}.} It remains to find a suitable threshold $\alpha$ such that characters with $\hat{y_i} \geq \alpha$ are considered a sentence boundary. In the simplest case, $\alpha$ can be set to a small constant value such as 0.01, where a higher $\alpha$ gives rise to more conservative segmentation. We denote the model variants with constant threshold as \ourunsup{}.
\ourunsup{} models can segment text according to the general Definition D1. In practice, models are trained on different corpora following different definitions of what makes up a sentence. This can be addressed to some extent by selecting the threshold $\alpha$ to maximise performance on the target corpus (\ourthresholdtuned{}). To allow for more sophisticated adaptation, we introduce an auxiliary objective.

\subsection{Auxiliary Punctuation-Prediction}
\label{ss:aux}

As an optional auxiliary objective, we predict the likelihood for punctuation characters among a predefined set $P$ to follow any character in the input text.\footnote{$P$ can be, e.g., the union of the $n$ most common punctuation characters in every language.} We remove characters among $P$ from the text with probability $p$. Let $\bm{p} \sim \text{Bernoulli}(p)^{|\bm{c}|}$ be a random binary mask. $\bm{x}'$ is corrupted in the same way as $\bm{x}$, with additional stochastic removal of punctuation characters among $P$. We \textbf{\color{ourred}highlight} the additional criterion over Equation~\eqref{equation:corruption} in the following Equation~\eqref{equation:aux_corruption}.
\begin{equation}
\bm{x}' = \left\{
c_i \:|\:
\begin{array}{l}
c_i \in \bm{c}, c_i \neq \mathtt{\backslash n},\\
\color{ourred}{\mathbf{c_i \notin P \text{ or } p_i = 0}}\\
\end{array}
\right\}
\label{equation:aux_corruption}
\end{equation}
In addition, we never remove two consecutive characters in Equation \eqref{equation:aux_corruption} to avoid ambiguity.\footnote{This constraint allows reconstructing the original sequence (modulo repetitions) from $\bm{x}$, $\bm{y}$ and $\bm{z}$. This would be prevented by removing two consecutive characters since $\bm{y}$ and $\bm{z}$ can only represent insertion of a single character.} The auxiliary labels $\bm{z}$ indicate, for the remaining characters, which (if any) character among $P$ followed them in the original sequence.
\begin{equation}
\bm{z} = \left\{\!
\begin{array}{l}
\!c_{i+1} \text{\; if $c_{i+1} \in P$}\\
\!0 \text{\;\;\;\;\;\;otherwise}
\end{array}
\:|\: c_i \in \bm{x}'
\right\}
\end{equation}
\begin{figure}[!t]
\centering
\includegraphics[width=0.99\linewidth]{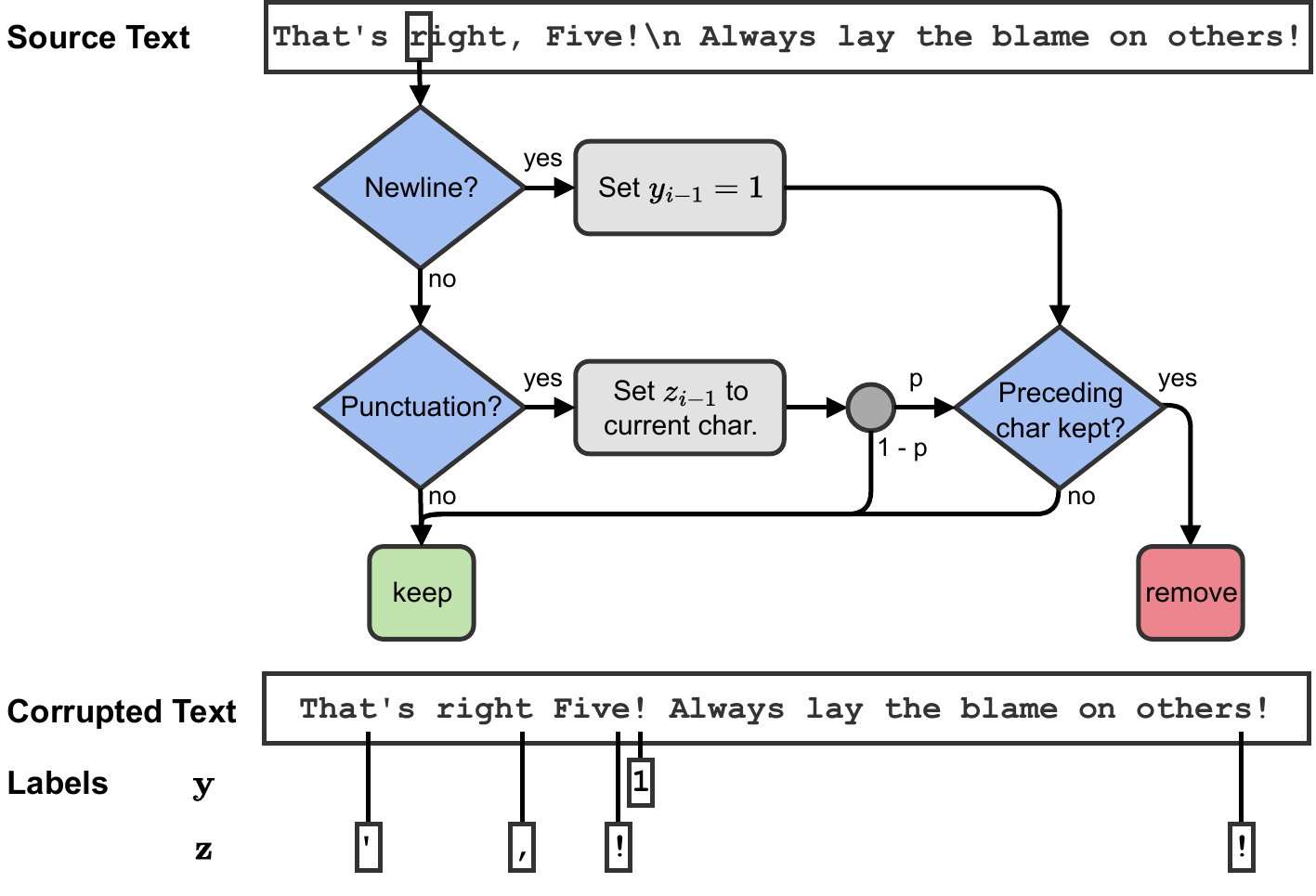}
\caption{Flow diagram of the corruption process (\S\ref{ss:aux}).}
\label{figure:flow}
\end{figure}
If the auxiliary objective is used, $\bm{y}$ and $\bm{h}$ are obtained using $\bm{x}'$ instead of $\bm{x}$ such that the same contextualized representations can be used for both objectives. Given predictions $\bm{\hat{z}}=\text{softmax}(q_\theta(\bm{h}))$, $\bm{\hat{z}} \in \mathbbm{R}^{|\bm{x}'|\times|P| + 1}$ where $q_\theta$ is an auxiliary prediction head parameterized by $\theta$, the auxiliary loss is defined as the categorical cross-entropy between $\bm{z}$ and $\bm{\hat{z}}$ as follows:
\begin{equation}
\mathcal{L}_{\theta}^{\mathrm{aux}} = -\frac{1}{|\bm{z}|} \sum_{i=0}^{|\bm{z}| - 1} \sum_{j = 0}^{|P|}\log \hat{z}_{i,j} \cdot \mathbb{I}(\mathrm{i}(z_i) = j).
\end{equation}
Here, $i$ assigns a unique index to every element of $P\cup\{0\}$. The total loss $\mathcal{L}_{\theta}$ is the sum of the primary objective of predicting newlines and the auxiliary objective of predicting punctuation.
\begin{equation}
\mathcal{L}_{\theta} = \mathcal{L}_{\theta}^{\mathrm{main}} + \mathcal{L}_{\theta}^{\mathrm{aux}} 
\label{equation:total_loss}
\end{equation}
Figure~\ref{figure:flow} summarizes the corruption process.
Sentence segmentation can be adapted to a target corpus with characters $\bm{c}_s$ and sentence boundaries $\bm{y}_s$ by finding the optimal coefficients $\bm{a}^* \in \mathbbm{R}^{|P|+1}$ of a logistic regression over punctuation logits, denoted as $h_{\bm{a}}(\bm{c}) = \text{sigmoid}(q_\theta(f_\theta(\bm{c})) \bm{a})$.
\begin{equation}
\bm{a}^* = \text{argmin}_{\bm{a}} \left\| 
\begin{array}{l}
y_s \odot \log h_{\bm{a}}(\bm{c}_s)\;+\\
(1 - y_s) \odot (1 - \log h_{\bm{a}}(\bm{c}_s)) 
\end{array}
\right\|
\end{equation}
This formulation is motivated by the hypothesis that the position of sentence boundaries is primarily determined by the probability distribution over punctuation marks at that position. It is a convex optimization problem. Effectively, it fits a one-layer neural network parameterized by $\bm{a}$ on the punctuation logits $q_\theta(f_\theta(\bm{c}))$ to predict sentence boundaries. We denote models adapted to a corpus using this method as \ourpuncttuned{}. We show how it leads to data-efficient adaptation later in \S\ref{sec:fewshot}.

\section{Experimental Setup}
\label{sec:experiments}

\subsection{Training Setup}
\label{sec:training}

We train a multilingual character-level language model on text in 85 languages. We sample text from all languages uniformly from the mC4 corpus \citep{2019t5}. In languages which use sentence-ending punctuation (every language besides Thai), we sample paragraphs such that a maximum of 10\% of paragraphs do not end in punctuation. For a list of languages, see Appendix~\ref{appendix:all_languages}.

We use the pretrained \canines{} ChLM \citep{clark-etal-2022-canine} as the starting point. To make the model sufficiently fast for sentence segmentation, we remove layers 4-12 from \canines{}, resulting in a 3-layer model (we also experiment with larger sizes in \S\ref{sec:results}). We add language adapters as described in \citet{pfeiffer-etal-2022-lifting} to efficiently increase per-language capacity in a modular fashion while keeping the same underlying model.

We continue training from the original \canines{} ChLM using our objective for 400k training steps with 512 characters per example and a batch size of 512.\footnote{This is character-wise equivalent to \textapprox10\% of \canines{} pretraining. It takes \textapprox 3 days on one TPUv3 with 8 cores.} We warm up the randomly initialized language adapters for 5k steps with a constant learning rate of $1$e$-4$ (keeping other parameters frozen), then start training the entire model with a linearly increasing learning rate from zero to $1$e$-4$ for the next 5k steps.\footnote{The learning rate was selected from the set \{$1$e$-4$, $5$e$-5$, $1$e$-5$\} to minimize loss on a held-out set of mC4. The learning rate schedule was not tuned.} We linearly decay the learning rate to zero over the remaining 390k steps. We use the AdamW optimizer \citep{DBLP:conf/iclr/LoshchilovH19}.

For the auxiliary objective, we choose $P$ to be the union of the 30 most common punctuation characters in every language (125 characters in total), and the removal probability $p=0.5$.

\subsection{Evaluation}
\label{sec:evaluation}

To evaluate our method, we determine how closely our sentence segmentation corresponds to the segmentation in different sentence-segmented corpora. Although the segmentation may vary due to what annotators of different corpora consider as a standalone sentence, it should match in the many cases which are largely unambiguous. We obtain sentence-segmented corpora from three different sources as follows.

\subrparagraph{1. Universal Dependencies} 
\citep[UD;][]{de-marneffe-etal-2021-universal, nivre-etal-2020-universal} is a collection of datasets annotating grammar (POS, morphological features and syntactic dependencies) in many different languages. Used UDv2.10 treebanks include segmentation into words and sentences.

\subrparagraph{2. OPUS100}
~\citep{zhang-etal-2020-improving} is a collection of sentences from various sources including subtitles and news sampled from OPUS~\citep{tiedemann-2012-parallel}. It consists of sentences in 100 languages with corresponding parallel sentences in English. Sentences in OPUS-100 sometimes lack proper punctuation, making it a challenging benchmark for sentence segmentation~\citep{sadvilkar-neumann-2020-pysbd}.

\subrparagraph{3. Ersatz}~\citep{wicks-post-2021-unified} introduces a collection of sentence-segmented text along with their sentence segmentation tool. The corpus consists primarily of sentences from the WMT shared tasks \citep{wmt-2020-machine,ws-2019-machine-translation} with manual corrections by the authors. Contrary to \citet{wicks-post-2021-unified}, we do not remove sentences without sentence-ending punctuation.

Evaluation details are shown in Appendix \ref{appendix:all_languages}. We use the test sets for evaluation. For adaptation via \ourthresholdtuned{} and \ourpuncttuned{}, we use the training sets. We evaluate by F1 score, where a character belongs to the positive class if it is followed by a sentence boundary. We set $\alpha = 0.01$ for \ourunsup{}.\footnote{The threshold was selected using the English UD data, so it is purely self-supervised for every language besides English.}

\rparagraph{Baselines} We compare against \spacysent{} and PySBD as representatives of rule-based methods, \spacydp{} as supervised punctuation-agnostic method, and Ersatz as a recent Transformer-based method; see \S\ref{sec:rw} again for their brief descriptions.

\rparagraph{Languages} For clarity, in the main paper we present results on Arabic, Czech, German, English, Spanish, Finnish, Hindi, Japanese, Georgian, Latvian, Polish, Thai, Xhosa and Chinese as a linguistically diverse subset ranging from low-resource (Georgian, Xhosa) to high-resource (German, English) languages. For the full evaluation in the remaining languages, see Appendix~\ref{appendix:all_languages}.

\section{Results and Discussion}
\label{sec:results}
\begin{table*}[!t]
\centering
\def\arraystretch{0.99}
\small
\setlength\tabcolsep{4pt}
\resizebox{\textwidth}{!}{
\begin{tabularx}{\linewidth}{llXXXXXXXXXXXXXX}
\toprule
& & \thead{ar} & \thead{cs} & \thead{de} & \thead{en} & \thead{es} & \thead{fi} & \thead{hi} & \thead{ja} & \thead{ka} & \thead{lv} & \thead{pl} & \thead{th} & \thead{xh} & \thead{zh}\\
\midrule
\multirow{8}{*}{UD} & \spacysent{} & 73.4 & 83.7 & 89.9 & 89.3 & 89.3 & 87.3 & 95.2 & 96.3 & - & 92.4 & 93.0 & - & - & 96.7 \\
& PySBD & 29.5 & - & 79.9 & 75.5 & 46.2 & - & 99.7 & 97.9 & - & - & 85.0 & - & - & 98.9 \\
& \spacydp{} & - & - & \textbf{96.9} & 91.7 & 99.0 & 95.5 & - & \textbf{98.0} & - & - & 98.5 & - & - & 98.2 \\
& Ersatz & 81.0 & 89.5 & 92.4 & 89.4 & 97.5 & 92.9 & 99.5 & 93.4 & - & 96.8 & 97.5 & - & - & 88.8 \\
& Punkt & - & 89.2 & 92.6 & 91.2 & 98.6 & 92.8 & - & - & - & - & 97.4 & - & - & - \\
\cmidrule{2-16}
& \ourunsup{} & 82.1 & 92.5 & 95.7 & 95.0 & 96.7 & 92.8 & 96.5 & 93.9 & - & 96.5 & 94.9 & \textbf{69.5} & - & 98.1 \\
& \ourthresholdtuned{} & 87.5 & 92.6 & 95.8 & 95.0 & 97.1 & 93.0 & 97.1 & 95.8 & - & 96.4 & 95.8 & - & - & 98.2 \\
& \ourpuncttuned{} & \textbf{88.2} & \textbf{95.5} & 96.7 & \textbf{96.7} & \textbf{99.7} & \textbf{98.2} & \textbf{99.9} & \textbf{98.0} & - & \textbf{99.1} & \textbf{99.4} & - & - & \textbf{99.8} \\
\midrule
\multirow{8}{*}{OPUS100} & \spacysent{} & 51.4 & 84.6 & 70.0 & 86.8 & 78.4 & 91.4 & 54.0 & 43.6 & - & 58.0 & 89.4 & - & - & 64.1 \\
& PySBD & 39.1 & - & 66.6 & 59.8 & 68.0 & - & 23.1 & 42.9 & - & - & 17.6 & - & - & 69.8 \\
& \spacydp{} & - & - & 74.5 & 89.4 & 88.4 & 92.9 & - & 42.2 & - & - & 92.9 & - & - & 69.4 \\
& Ersatz & 59.7 & 86.2 & 73.2 & 87.7 & 90.0 & 92.9 & 58.5 & 28.3 & - & 77.6 & 92.2 & - & - & 54.7 \\
& Punkt & - & 86.5 & 73.5 & 88.6 & 90.2 & 93.5 & - & - & - & - & 92.8 & - & - & - \\
\cmidrule{2-16}
& \ourunsup{} & 66.2 & 88.5 & 78.5 & 91.3 & 90.8 & 91.5 & 66.7 & 44.9 & 91.9 & 79.6 & 92.4 & 68.8 & 78.7 & 81.0 \\
& \ourthresholdtuned{} & 66.4 & 90.8 & 85.8 & 90.3 & 92.1 & 93.1 & 66.1 & 80.5 & 91.7 & 86.5 & 92.8 & 71.5 & 81.9 & 77.8 \\
& \ourpuncttuned{} & \textbf{77.2} & \textbf{95.2} & \textbf{90.1} & \textbf{95.0} & \textbf{95.4} & \textbf{96.1} & \textbf{77.5} & \textbf{87.4} & \textbf{93.2} & \textbf{91.9} & \textbf{96.0} & \textbf{72.9} & \textbf{90.4} & \textbf{89.2} \\
\midrule
\multirow{8}{*}{Ersatz} & \spacysent{} & 89.4 & 84.1 & 89.9 & 89.8 & 85.0 & 94.7 & 89.9 & 84.7 & - & 89.8 & 77.6 & - & - & 90.6 \\
& PySBD & 47.9 & - & 95.5 & 74.2 & 84.6 & - & 87.8 & 87.7 & - & - & 46.1 & - & - & 92.7 \\
& \spacydp{} & - & - & 96.3 & 98.3 & 96.4 & 95.2 & - & 91.2 & - & - & 94.4 & - & - & 95.8 \\
& Ersatz & 92.9 & 96.8 & 95.6 & 97.6 & 96.7 & 95.9 & \textbf{96.9} & 85.9 & - & 98.7 & 95.1 & - & - & 87.6 \\
& Punkt & - & 96.8 & 95.5 & 97.8 & 96.6 & 95.7 & - & - & - & - & 94.3 & - & - & - \\
\cmidrule{2-16}
& \ourunsup{} & 87.8 & 93.7 & 95.7 & 96.8 & 98.8 & 97.5 & 94.4 & 81.5 & - & 97.2 & 94.8 & - & - & 93.5 \\
& \ourthresholdtuned{} & 88.9 & 94.1 & 96.0 & 96.9 & 97.8 & 97.3 & 94.7 & 82.6 & - & 97.3 & 92.8 & - & - & 93.7 \\
& \ourpuncttuned{} & \textbf{92.9} & \textbf{98.9} & \textbf{99.3} & \textbf{98.7} & \textbf{99.5} & \textbf{99.4} & 96.4 & \textbf{94.8} & - & \textbf{99.4} & \textbf{98.0} & - & - & \textbf{97.8} \\
\bottomrule
\end{tabularx}
}
\caption{Sentence segmentation F1 scores. For Georgian (ka), Thai (th), and Xhosa (xh), no Ersatz and UD corpora are available. For Thai, no OPUS100 training data is available. We adapt \ourthresholdtuned{} and \ourpuncttuned{} to each corpus using the corresponding training datasets. Bold numbers indicate the best results for each language and dataset.}
\label{table:comparison}
\end{table*}

\begin{figure*}[t]
\centering
\hspace*{-2mm}
\includegraphics[width=1.02\textwidth]{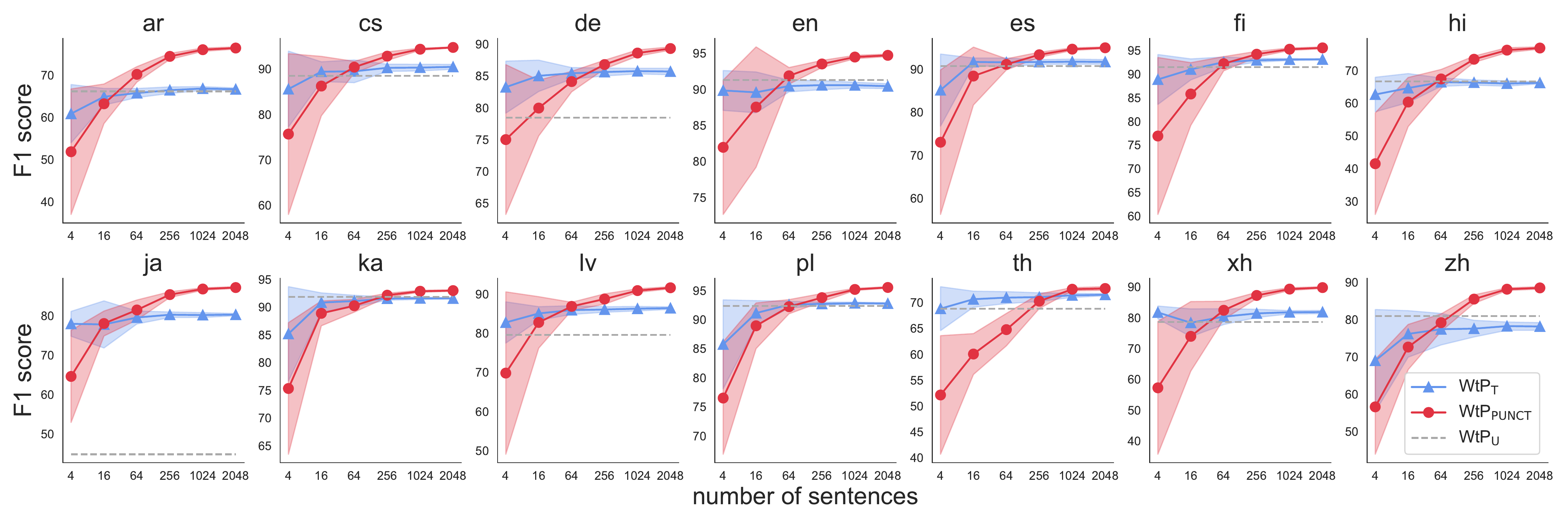}
\caption{F1 score vs. number of sentences used for adaptation on the OPUS100 datasets. Numbers are averaged over 20 different sentence samples of the respective sizes. Shaded regions indicate one standard deviation.}
\label{figure:fewshot}
\end{figure*}

Main results are shown in Table~\ref{table:comparison}, results for all languages in Appendix \ref{appendix:all_languages}.
For UD and Ersatz, \ourunsup{}, though self-supervised, matches the performance of the best prior tool (usually \spacydp{} or Ersatz) on most languages. It falls behind on Hindi (hi), Japanese (ja) and Chinese (zh). While the majority of the languages we train on uses Latin punctuation characters, Hindi, Japanese and Chinese use punctuation in their own script, which could cause this deficit. In addition, Japanese and Chinese do not use whitespace between words and sentences which makes segmentation more challenging.
The deficit can be resolved by supervised adaptation: \ourpuncttuned{} surpasses the prior best on 19 out of 22 of the UD and Ersatz datasets (2.1\% on average), including 4 out of 6 of the Hindi, Japanese and Chinese datasets. On OPUS100, \ourunsup{} already outperforms the best prior tool in 10 out of 12 languages (3.2\% on average). Adaptation via \ourthresholdtuned{} and \ourpuncttuned{} increases the gap to 7.8\% and 14.1\%, respectively. The strong increase in performance on OPUS100 compared to prior tools may be caused by \ourmethod{} being pretrained on web text, which could be better suited for the generally noisy text in OPUS100.

\begin{table}[!t]
\footnotesize
\centering
\setlength\tabcolsep{5pt}
\begin{tabularx}{\linewidth}{lccc|c}
\toprule
& \thead{ORCHID} & \thead{UD} & \thead{OPUS100} & \textbf{\textit{\thead{Macro\\Avg.}}}\\
\midrule
PyThaiNLP & 55.6 & 64.7 & 62.5 & 60.9\\
\midrule
\ourunsup{} & 67.9 & 69.5 & 68.8 & 68.7\\
\ourthresholdtuned{}$_\textsc{:opus100}$ & \textbf{69.2} & \textbf{78.0} & 71.5 & \textbf{72.9}\\
\ourpuncttuned{}$_\textsc{:opus100}$ & 51.8 & 77.1 & \textbf{72.9} & 67.3\\
\bottomrule
\end{tabularx}
\caption{Thai segmentation F1 scores. Score on full dataset for ORCHID, on test sets for UD and OPUS100. \ourmethod{}$_{*\textsc{:opus100}}$ denotes \ourmethod{} adapted to the OPUS100 corpus; we do not use ORCHID and UD for adaptation.}
\label{table:thai}
\end{table}

\rparagraph{Punctuation-Free Segmentation in Thai}
\label{sec:thai}
Thai is especially relevant to our method since it is the only language in our set which does not, in general, use punctuation to separate sentences. Most prior sentence segmentation tools rely on punctuation \citep{sadvilkar-neumann-2020-pysbd, wicks-post-2021-unified, kiss-strunk-2006-unsupervised}, which has led to the development of Thai-specific segmentation tools in a separate strand of work \citep{saetia2019semi, nararatwong2018improving, zhou-etal-2016-word, pythainlp}. Our method is the first to segment Thai and other languages using the same methodology. To verify \ourmethod{} performs as expected on Thai, we evaluate it on the Thai ORCHID corpus \citep{sornlertlamvanich1997orchid} in addition to UD and OPUS100, and compare against the Open-Source PyThaiNLP toolkit's sentence segmentation \citep{pythainlp}: PyThaiNLP v3.1.1.. Results are shown in Table \ref{table:thai}. While scores are lower than for most other languages due to the more challenging nature of sentence segmentation in Thai, \ourmethod{} consistently outperforms PyThaiNLP. We also find that transfer from \ourpuncttuned{} tuned on OPUS100 to ORCHID and UD is limited; see Appendix \ref{appendix:transfer}.

\rparagraph{Few-Shot Adaptation} 
\label{sec:fewshot}
We now analyse how many sentences are needed to successfully adapt \ourmethod{} to a target corpus. Since we observed the most substantial need for adaptation on OPUS100, we focus on the OPUS100 datasets. Figure~\ref{figure:fewshot} shows performance of \ourthresholdtuned{} and \ourpuncttuned{} w.r.t. the number of sentences used for adaptation. Between 64 and 256 sentences, \ourpuncttuned{} consistently starts outperforming the self-supervised \ourunsup{} baseline and \ourthresholdtuned{} in all languages except Thai.\footnote{Assuming annotation speed of \textapprox4 sentences per minute, it would take 15 minutes to 1 hour to annotate 64-256 sentences.} It still benefits from more data up to \textapprox 2k sentences. Using less than 64 sentences, 
\ourthresholdtuned{} outperforms \ourpuncttuned{}, but in some cases does not surpass \ourunsup{}. In Thai, adaptation via \ourpuncttuned{} is less effective. However, with enough data, it can still improve over \ourthresholdtuned{}.

\begin{table}[!t]
\footnotesize
\def\arraystretch{1.0}
\centering
\begin{tabularx}{\linewidth}{cc r rrrr}
\toprule
\thead{src} & \thead{tgt} & \thead{True} & \thead{None} & \thead{Naïve} & \thead{Ersatz} & \hspace*{-0.2cm}\thead{\ourpuncttuned{}}\\
\cmidrule(lr){1-2} \cmidrule(lr){3-3} \cmidrule(lr){4-7} 
ar & en & 44.0 & 16.4 & 29.4 & 38.0 & \textbf{40.8}\\
cs & en & 37.7 & 30.6 & 25.5 & 36.6 & \textbf{37.6}\\
de & en & 37.5 & 31.8 & 31.5 & 36.8 & \textbf{37.2}\\
en & eo & 48.8 & 21.8 & 26.7 & 44.8 & \textbf{47.2}\\
es & en & 43.7 & 37.3 & 33.5 & 43.1 & \textbf{43.2}\\
fi & en & 28.9 & 20.8 & 20.1 & 28.3 & \textbf{28.8}\\
hi & en & 39.2 & 8.7 & 21.0 & 28.1 & \textbf{32.7}\\
ja & en & 18.6 & 3.2 & 9.0 & 7.9 & \textbf{16.2}\\
ka & en & 22.3 & 0.6 & 11.6 & - & \textbf{21.5}\\
lv & en & 54.1 & 30.3 & 43.4 & 51.3 & \textbf{53.6}\\
pl & en & 29.1 & 22.4 & 18.7 & 28.5 & \textbf{29.0}\\
th & en & 27.1 & 4.4 & 21.4 & - & \textbf{22.6}\\
xh & en & 61.2 & 17.4 & 31.7 & - & \textbf{56.8}\\
zh & en & 48.4 & 27.4 & 35.6 & 44.8 & \textbf{47.0}\\
\cmidrule(lr){1-2} \cmidrule(lr){3-3} \cmidrule(lr){4-7} 
\multicolumn{2}{c}{\textbf{\textit{Macro Avg.}}} & 39.1 & 22.8 & 26.7 & 35.3 & \textbf{37.6}\\
\bottomrule
\end{tabularx}
\caption{Impact of sentence segmentation on machine translation BLEU score. \textit{True} indicates the ground truth segmentation.
Average excludes languages with missing Ersatz scores. English is translated to Esperanto (eo).}
\label{table:extrinsic}
\end{table}

\rparagraph{Downstream Impact of Sentence Segmentation}
\label{sec:downstream}
Prior sentence segmentation tools limit themselves to intrinsic performance evaluation \citep{gillick-2009-sentence, sadvilkar-neumann-2020-pysbd, wicks-post-2021-unified}.
But does proper sentence segmentation actually matter for downstream tasks?
Usually sentence segmentation is done as the first step in a pipeline i.e. text is first split into sentences, then every sentence is passed one-by-one through some model to solve a downstream task. We quantify the impact of sentence segmentation on one downstream task: machine translation (MT). We obtain MT models for 14 languages trained on OPUS100 data from \mbox{OPUS-MT}~\citep{tiedemann-thottingal-2020-opus}. We simulate a real-world scenario by partitioning the OPUS100 test data into paragraphs consisting of 10 sentences each. We sentence-segment the paragraphs using different tools, pass every sentence through the MT model, and conjoin the resulting translation. In addition, we compare against two baselines: (i) \textit{None}, where the entire paragraph is passed to the MT model at once and (ii) \textit{Naïve}, where the paragraph is segmented into 10 equally long sequences of words or characters.\footnote{Words if segmentation into words is trivial (i.e. the language uses whitespace between words), otherwise characters.}

We evaluate the predicted translation via BLEU score~\citep{papineni-etal-2002-bleu} implemented in SacreBLEU~\citep{post-2018-call}.\footnote{nrefs:1|case:mixed|eff:no|tok:13a|smooth:exp|version:2.2.1} Results are shown in Table \ref{table:extrinsic}. We find that sentence segmentation is necessary for models trained on the sentence-level: Passing the entire paragraph to the model at once causes a drop of 16.3 BLEU score compared to the ground truth segmentation. Furthermore, the choice of sentence segmentation tool makes a difference: \ourpuncttuned{} outperforms Ersatz by an average 2.3 BLEU points.

Previous work has found no clear difference between Ersatz, Moses, Punkt and \spacysent{} for German-English Machine Translation on OPUS \citep{wicks-post:2022:WMT}. We find that, although German-English translation exhibits one of the lowest differences, using \ourmethod{} to match segmentation to the segmentation used during training can lead to consistent and sometimes large improvements across all languages in our evaluation. For qualitative analysis, see Appendix \ref{appendix:mt_examples}.

\begin{table}[!t]
\footnotesize
\def\arraystretch{1.0}
\centering
\setlength\tabcolsep{3pt}
\begin{tabularx}{\linewidth}{Xrrrrrr|c}
\toprule
& \multicolumn{3}{c}{\thead{English}} & \multicolumn{3}{c|}{\thead{Bengali}} & \multirow{2}{*}[0.5em]{\textbf{\textit{\thead{Macro\\Avg.}}}}\\
& C & P & Q & C & P & Q &\\
\midrule
\canines{} & 66.6 & 80.2 & 72.3 & 35.7 & 61.2 & 29.9 & 57.7\\
\midrule
\ourpuncttuned{} & 61.3 & 76.0 & 73.3 & \textbf{38.9} & \textbf{72.8} & 39.2 & 60.3\\
\ourmethod{}$_\textsc{Finetune}$ & \textbf{69.8} & \textbf{82.7} & \textbf{77.9} & 36.6 & 72.5 & \textbf{40.9} & \textbf{63.4}\\
\bottomrule
\end{tabularx}
\caption{Punctuation restoration F1 score across \textit{comma} (C), \textit{period} (P) and \textit{question mark} (Q) of \canines{}, \ourpuncttuned{} and the fully finetuned \ourmethod{} model (\ourmethod{}$_\textsc{Finetune}$). Train and test splits as in \citet{alam-etal-2020-punctuation}. Details in Appendix \ref{appendix:punctuation_restoration}.}
\label{table:punctuation_restoration}
\end{table}

\rparagraph{Application to the Punctuation Restoration Task}
\label{sec:punctuation_restoration}
Our auxiliary punctuation-prediction objective is closely related to the task of punctuation restoration \citep{puaics2022capitalization}, where the position of punctuation characters is predicted in unpunctuated text. We evaluate the capacity of \ourmethod{} for punctuation restoration on the English IWSLT dataset \citep{CHE16.103, cettolo2013report}, as well as a Bengali dataset provided by \citet{alam-etal-2020-punctuation}. We compare against the pretrained \canines{} (the starting point for the \ourmethod{} models) fine-tuned on punctuation restoration data in the respective language. Results are shown in Table~\ref{table:punctuation_restoration}. \canines{} with continued training via \ourmethod{} outperforms the off-the-shelf \canines{} by an average 5.7\% F1 points.\footnote{For a fair comparison, we use the 12-layer \ourmethod{} model.} Fitting a logistic regression on punctuation logits (\ourpuncttuned{}) outperforms the fully fine-tuned \canines{} on Bengali, but not on English.
While prior work approaches punctuation restoration by corrupting a curated small-scale corpus \citep{nguyen2019fast, ueffing2013improved}, we are the first to show that Web-scale pretraining via corruption can improve punctuation restoration.

\rparagraph{Ablation Studies}
\label{sec:ablations}
We now quantify the impact of multiple design choices of \ourmethod{} in Table~\ref{table:ablations}. \textit{No language adapters} expectedly decreases performance. We remove language adapters by using \citet{pfeiffer-etal-2022-lifting}'s $\textsc{Shared}$ setting (keeping the language adapter layers but sharing parameters between languages) to match FLOPs. \textit{No punctuation-specific sampling} i.e. not sampling paragraphs in punctuated languages such that a maximum of 10\% do not end in punctuation decreases performance, especially of \ourunsup{}. Removing the auxiliary objective (\textit{No aux. punctuation-prediction}) does not allow adaptation via \ourpuncttuned{}, but also decreases performance of \ourunsup{} and \ourthresholdtuned{}. Corrupting punctuation, but not adding the auxiliary loss in Equation~\eqref{equation:total_loss} (\textit{No aux. punctuation-prediction + punctuation corruption}) also decreases performance. This implies that, besides enabling adaptation via \ourpuncttuned{}, the auxiliary punctuation-prediction objective positively impacts the main newline-prediction objective.

Reducing the amount of pretraining data has comparatively little impact on performance, possibly because newline- and punctuation-prediction are not particularly knowledge-intensive tasks. Reducing the amount of layers to one decreases performance by a considerable amount. Scaling to 6, 9 and 12 layers continues to improve performance.

\begin{table}[t!]
\footnotesize
\centering
\setlength\tabcolsep{0pt}
\begin{tabularx}{\linewidth}{lrrr}
\toprule
\thead{Variation} & \ourunsup$\quad$ & \ourthresholdtuned{}$\quad$ & \ourpuncttuned{}\\
\midrule
Original & 89.0 & 90.4 & 94.9\\
\midrule
No language adapters & 89.0 & 90.3 & 94.6\\
No punctuation-specific sampling & 87.9 & 90.5 & 94.8\\
No aux. punctuation-prediction & 88.3 & 90.0 & -\\
\quad+ punctuation corruption & 88.4 & 89.8 & -\\
\midrule
Reduced pretraining data size\\
\quad75\% subsample & 89.1 & 90.6 & 94.9\\
\quad50\% subsample & 89.2 & 90.6 & 94.8\\
\quad25\% subsample & 89.1 & 90.4 & 94.8\\
\midrule
Scaled amount of layers\\
\quad\phantom{0}1 layer & 88.8 & 90.1 & 94.4\\
\quad\phantom{0}3 layers (Original) & 89.0 & 90.4 & 94.9\\
\quad\phantom{0}6 layers & 89.3 & 90.8 & 95.1\\
\quad\phantom{0}9 layers & 89.8 & 91.1 & 95.3\\
\quad12 layers & 89.9 & 91.2 & 95.5\\
\bottomrule
\end{tabularx}
\caption{Ablation studies and sensitivity analysis w.r.t. amount of layers and amount of pretraining data.}
\label{table:ablations}
\end{table}

\section{Conclusion}

We have introduced \ourmethod{}, a method for multilingual sentence segmentation without relying on punctuation or sentence-segmented training data. We have demonstrated strong performance of our self-supervised sentence segmentation method across 85 languages and 3 different corpora, matching performance of the best prior (supervised) tools. We have further improved \ourmethod{} in a data-efficient way by means of inexpensive supervised adaptation, which leads to state-of-the-art scores outperforming the best prior tools by an average 6.1\% F1 points. We have also found that, though often an overlooked component of NLP systems, sentence segmentation can have a strong impact on downstream tasks by showing that matching the segmentation at inference to the segmentation used during training benefits sentence-level MT models.

\section*{Limitations}

\ourmethod{} performs comparatively worse in some low-resource languages (e.g. Welsh, Nepalese, Punjabi, Pushto). This may be attributed to quality issues of mC4 in these languages \citep{kreutzer-etal-2022-quality}. In addition, we find that the adapted \ourpuncttuned{} classifiers generally do not transfer well across languages and dataset collections (Appendix~\ref{appendix:transfer}). Finally, although bias is less obvious in segmentation tasks than e.g. generation, \ourmethod{} may be biased by performing disproportionately well on text by communities which are overrepresented in the training data, while performing worse on text from underrepresented communities. We try to minimize this form of bias by sampling text from all languages uniformly.

\section*{Acknowledgements}
Ivan Vuli\'{c} is supported by a personal Royal Society University Research Fellowship \textit{`Inclusive and Sustainable Language Technology for a Truly Multilingual World'} (no 221137; 2022--).

Research supported with Cloud TPUs from Google's TPU Research Cloud (TRC). We thank Christoph Minixhofer for advice in the initial stage of this project.
We also thank Sebastian Ruder and Srini Narayanan  for helpful feedback on a draft of this paper. 

\bibliography{anthology,custom}
\bibliographystyle{acl_natbib}

\newpage

\appendix

\section{Results in All Languages}
\label{appendix:all_languages}

We give an overview over all used languages for pretraining in Table~\ref{table:list_of_languages}. Results of \ourmethod{} and prior methods for all languages are shown in Tables~\ref{table:all_comparisons_first}-\ref{table:all_comparisons_last}.

\section{Qualitative Analysis of the Impact on Machine Translation}
\label{appendix:mt_examples}

An example paragraph of the German-English OPUS100 data is shown in Table~\ref{table:mt_examples}. Passing the entire paragraph to the model at once (\textit{None}) misses a considerable portion of the latter parts of the input text. Semantically uninformed chunking (\textit{Naïve}) misses some parts where semantics are split across chunks. Segmentation with Ersatz results in longer sentences since text can not be split on non-punctuation characters, and adds a wrong boundary after 'p.'. Although \ourpuncttuned{} does not exactly match the segmentation in OPUS100 (e.g. the first sentence, which could be considered undersegmented in the ground truth, is split into three parts), it identifies all correct sentence boundaries.

\section{Punctuation Restoration Details}
\label{appendix:punctuation_restoration}

For \ourpuncttuned{}, we fit a logistic regression on the punctuation logits to predict one of four classes for each character (\textit{comma}, \textit{period}, \textit{question mark}, \textit{none}). For \ourmethod{}$_\textsc{Finetune}$, we replace the pretrained prediction head $g_\theta$ with a new 4-class prediction head and train the entire model with a batch size of 32 for 5 epochs at 256 characters sequence length. We use the AdamW optimizer with a triangular learning rate schedule peaking at 1e-5 at 30\% of training steps. For \canines{}, we again add a 4-class prediction head on top of the pretrained model and use the same hyperparameters as for \ourmethod{}$_\textsc{Finetune}$. Hyperparameters were chosen by setting them to reasonable defaults (no hyperparameter tuning).

\section{Transferring \ourmethod{} across Languages and Collections}
\label{appendix:transfer}

\begin{figure*}[t!]
\centering
\hspace*{-1.3cm}
\includegraphics[width=18cm]{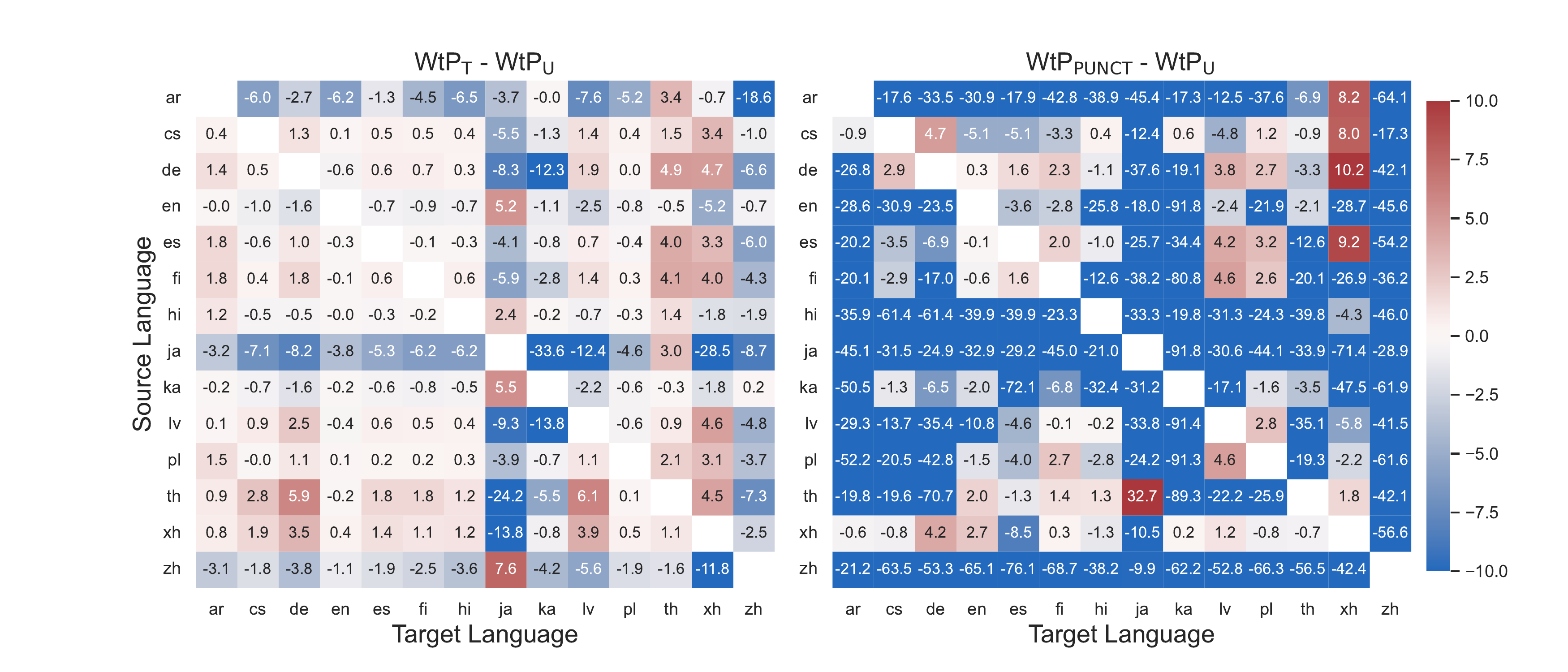}
\caption{Performance of cross-lingual transfer. Values indicate the average difference of the adaptation method to the self-supervised \ourunsup{} baseline when trained on data in the source language and evaluated on data in the target language. We average scores across corpora, and use the same corpus collection for training and evaluation.}
\label{figure:crosslingual}
\end{figure*}

\begin{figure*}[t!]
\centering
\hspace*{-1.3cm}
\includegraphics[width=18cm]{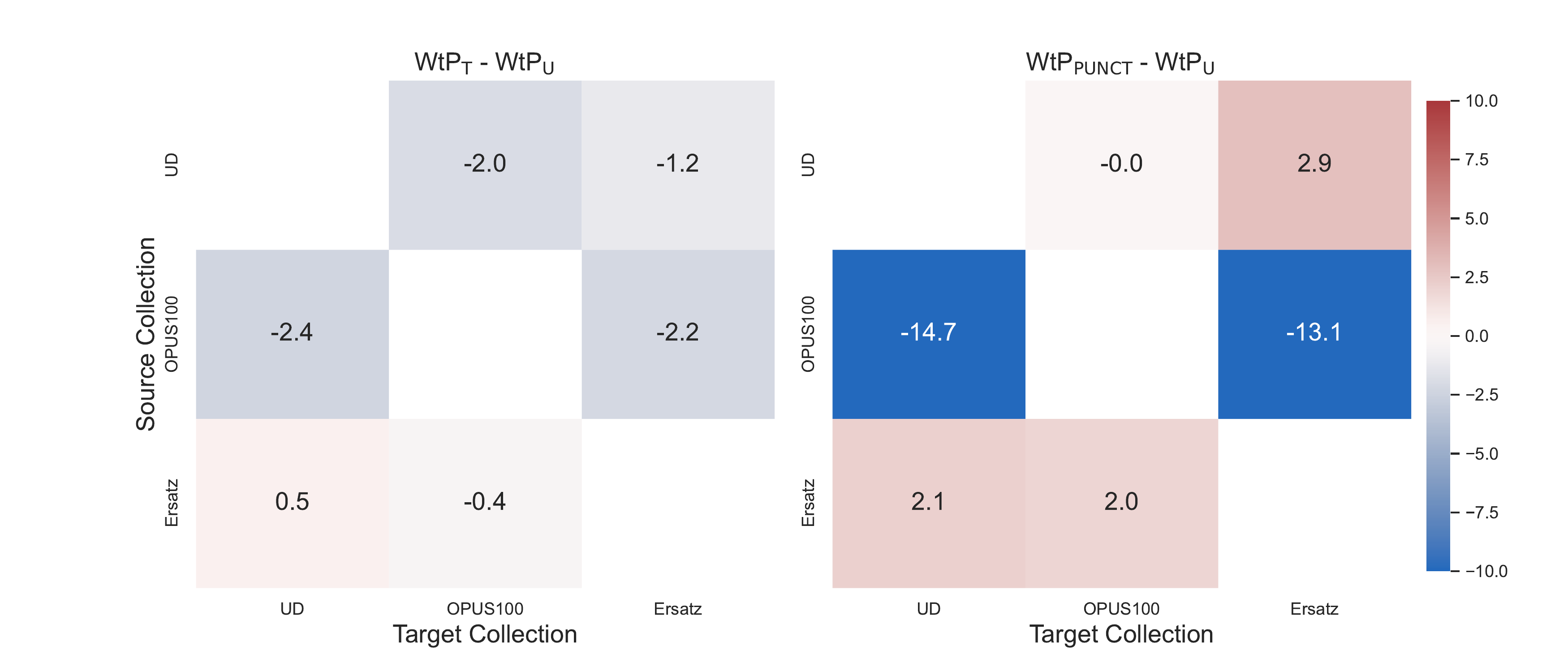}
\caption{Performance of cross-collection transfer. Values indicate the average difference of the adaptation method to the self-supervised \ourunsup{} baseline when trained on data in the source collection and evaluated on data in the target collection. Scores are averaged across languages.}
\label{figure:crosscollection}
\end{figure*}

We investigate the capacity of \ourmethod{} models for cross-lingual transfer and transfer across collections by training \ourthresholdtuned{} and \ourpuncttuned{} using supervised data in one corpus, then evaluating performance on a different corpus and comparing against the \ourunsup{} baseline. For cross-lingual transfer, we keep the collection the same between source and target; for cross-collection transfer, we keep the language the same. Results for cross-lingual transfer are shown in Figure \ref{figure:crosslingual}, cross-collection transfer in Figure \ref{figure:crosscollection}. We find that the threshold estimated by \ourthresholdtuned{} can generally be transferred. This is consistent with our observation that \ourunsup{} performs well in many languages although the threshold for \ourunsup{} was selected using only one corpus (English UD).
More sophisticated adaptation via \ourpuncttuned{} generally hurts cross-lingual transfer, although some directions (e.g. th$\,\rightarrow\,$ja, es$\,\rightarrow\,$xh) exhibit strong positive transfer. Across collections, transferring from Ersatz and UD leads to moderate improvements, while transferring from OPUS100 strongly decreases performance.

\begin{table*}[t]
\centering
\footnotesize
\resizebox{\textwidth}{!}{%
\begin{tabular}{llc|rrr|}
\toprule
Language & iso & Space & UD & OPUS100 & Ersatz \\
\midrule
Afrikaans & af &  & AfriBooms (425) & 1.9k & \\
Amharic & am &  &  & 2.0k & \\
Arabic & ar &  & PADT (680) & 2.0k & 1.5k\\
Azerbaijani & az &  &  & 2.0k & \\
Belarusian & be &  & HSE (1.1k) & 2.0k & \\
Bulgarian & bg &  & BTB (1.1k) & 2.0k & \\
Bengali & bn &  & BRU (56) & 2.0k & \\
Catalan & ca &  & AnCora (1.8k) & 2.0k & \\
Cebuano & ceb &  & GJA (188) &  & \\
Czech & cs &  & PDT (10.1k) & 2.0k & 1.7k\\
Welsh & cy &  & CCG (953) & 1.8k & \\
Danish & da &  & DDT (565) & 2.0k & \\
German & de &  & GSD (977) & 1.9k & 2.0k\\
Greek & el &  & GDT (456) & 2.0k & \\
English & en &  & GUM (1.1k) & 2.0k & 7.7k\\
Esperanto & eo &  &  & 2.0k & \\
Spanish & es &  & AnCora (1.7k) & 2.0k & 3.1k\\
Estonian & et &  & EDT (3.2k) & 2.0k & 2.0k\\
Basque & eu &  & BDT (1.8k) & 2.0k & \\
Persian & fa &  & PerDT (1.5k) & 2.0k & \\
Finnish & fi &  & TDT (1.6k) & 2.0k & 2.0k\\
French & fr &  & GSD (416) & 2.0k & 1.7k\\
Western Frisian & fy &  &  & 1.9k & \\
Irish & ga &  & IDT (454) & 2.0k & \\
Scottish Gaelic & gd &  & ARCOSG (545) & 1.1k & \\
Galician & gl &  & TreeGal (400) & 2.0k & \\
Gujarati & gu &  &  & 1.9k & 1.0k\\
Hausa & ha &  &  & 2.0k & \\
Hebrew & he &  & IAHLTwiki (393) & 2.0k & \\
Hindi & hi &  & HDTB (1.7k) & 2.0k & 2.5k\\
Hungarian & hu &  & Szeged (449) & 2.0k & \\
Armenian & hy &  & BSUT (595) & 7.0k & \\
Indonesian & id &  & PUD (1.0k) & 2.0k & \\
Igbo & ig &  &  & 1.7k & \\
Icelandic & is &  & IcePaHC (5.2k) & 2.0k & \\
Italian & it &  & ISDT (482) & 2.0k & \\
Japanese & ja & \ding{55} & GSD (543) & 2.0k & 1.1k\\
Javanese & jv &  & CSUI (125) &  & \\
Georgian & ka &  &  & 2.0k & \\
Kazakh & kk &  & KTB (1.0k) & 1.9k & 1.0k\\
Khmer & km & \ding{55} &  & 1.9k & 2.4k\\
Kannada & kn &  &  & 906 & \\
Korean & ko &  & Kaist (2.3k) & 2.0k & \\
\bottomrule
\end{tabular}

\begin{tabular}{|llc|rrr}
\toprule
Language & iso & Space & UD & OPUS100 & Ersatz \\
\midrule
Kurdish & ku &  &  & 1.9k & \\
Kirghiz & ky &  &  & 1.7k & \\
Latin & la &  & ITTB (2.1k) &  & \\
Lithuanian & lt &  & ALKSNIS (684) & 2.0k & 1.0k\\
Latvian & lv &  & LVTB (2.3k) & 2.0k & 2.0k\\
Malagasy & mg &  &  & 2.0k & \\
Macedonian & mk &  &  & 2.0k & \\
Malayalam & ml &  &  & 2.0k & \\
Mongolian & mn &  &  & 4.2k & \\
Marathi & mr &  & UFAL (47) & 2.0k & \\
Malay & ms &  &  & 1.9k & \\
Maltese & mt &  & MUDT (518) & 2.0k & \\
Burmese & my & \ding{55} &  & 2.0k & \\
Nepalese & ne &  &  & 1.9k & \\
Dutch & nl &  & Alpino (596) & 2.0k & \\
Norwegian & no &  & Bokmaal (1.9k) & 2.0k & \\
Panjabi & pa &  &  & 2.0k & \\
Polish & pl &  & PDB (2.2k) & 2.0k & 1.0k\\
Pushto & ps &  &  & 1.8k & 2.7k\\
Portuguese & pt &  & Bosque (1.2k) & 2.0k & \\
Romanian & ro &  & Nonstandard (1.1k) & 2.0k & 2.0k\\
Russian & ru &  & Taiga (881) & 2.0k & 991\\
Sinhala & si &  &  & 2.0k & \\
Slovak & sk &  & SNK (1.1k) & 2.0k & \\
Slovenian & sl &  & SSJ (1.3k) & 2.0k & \\
Albanian & sq &  & TSA (60) & 2.0k & \\
Serbian & sr &  & SET (520) & 2.0k & \\
Swedish & sv &  & LinES (1.0k) & 2.0k & \\
Tamil & ta &  & TTB (120) & 2.0k & 1.0k\\
Telugu & te &  &  & 2.0k & \\
Tajik & tg &  &  & 2.0k & \\
Thai & th &  & PUD (1.0k) & 2.0k & \\
Turkish & tr &  & IMST (983) & 2.0k & 3.0k\\
Ukrainian & uk &  & IU (892) & 2.0k & \\
Urdu & ur &  & UDTB (535) & 1.9k & \\
Uzbek & uz &  &  & 2.0k & \\
Vietnamese & vi &  & VTB (800) & 1.9k & \\
Xhosa & xh &  &  & 1.9k & \\
Yiddish & yi &  &  & 1.3k & \\
Yoruba & yo &  & YTB (318) & 9.4k & \\
Chinese & zh & \ding{55} & GSDSimp (500) & 2.0k & 2.0k\\
Zulu & zu &  &  & 1.9k & \\
& & & & &\\
\bottomrule
\end{tabular}
}
\caption{List of the 85 languages used in pretraining, whether they generally use whitespace to separate sentences, and their corresponding evaluation dataset sizes in sentences. For UD, the treebank name is also shown.}
\label{table:list_of_languages}
\end{table*}

\begin{table*}[t!]
\centering
\small
\setlength\tabcolsep{4pt}
\resizebox{\textwidth}{!}{
\begin{tabularx}{\linewidth}{llXXXXXXXXXXXXXXX}
\toprule
& & \thead{af} & \thead{am} & \thead{ar} & \thead{az} & \thead{be} & \thead{bg} & \thead{bn} & \thead{ca} & \thead{ceb} & \thead{cs} & \thead{cy} & \thead{da} & \thead{de} & \thead{el} & \thead{en}\\
\midrule
\multirow{8}{*}{UD} & \spacysent{} & 98.6 & - & 73.4 & - & - & 90.7 & \textbf{100.0} & 95.3 & - & 83.7 & - & 85.7 & 89.9 & 90.5 & 89.3 \\
& PySBD & - & - & 29.5 & - & - & 74.5 & - & - & - & - & - & 72.4 & 79.9 & 91.6 & 75.5 \\
& \spacydp{} & - & - & - & - & - & - & - & \textbf{99.8} & - & - & - & 94.7 & \textbf{96.9} & 94.4 & 91.7 \\
& Ersatz & - & - & 81.0 & - & - & - & - & - & - & 89.5 & - & - & 92.4 & - & 89.4 \\
& Punkt & - & - & - & - & - & - & - & - & - & 89.2 & - & 94.3 & 92.6 & 93.1 & 91.2 \\
\cmidrule{2-17}
& \ourunsup{} & 98.0 & - & 82.1 & - & 89.8 & 98.2 & 94.1 & 98.4 & \textbf{99.7} & 92.5 & 99.2 & 95.2 & 95.7 & 97.4 & 95.0 \\
& \ourthresholdtuned{} & 99.1 & - & 87.5 & - & 89.6 & 98.1 & - & 98.5 & - & 92.6 & 98.9 & 94.6 & 95.8 & \textbf{97.7} & 95.0 \\
& \ourpuncttuned{} & \textbf{99.9} & - & \textbf{88.2} & - & \textbf{92.1} & \textbf{99.6} & - & 99.8 & - & \textbf{95.5} & \textbf{99.5} & \textbf{98.6} & 96.7 & \textbf{97.7} & \textbf{96.7} \\
\midrule
\multirow{8}{*}{OPUS100} & \spacysent{} & 30.7 & 6.6 & 51.4 & 70.6 & - & 91.5 & 78.6 & 86.2 & - & 84.6 & - & 87.6 & 70.0 & 82.5 & 86.8 \\
& PySBD & - & 6.2 & 39.1 & - & - & 72.9 & - & - & - & - & - & 70.3 & 66.6 & 62.5 & 59.8 \\
& \spacydp{} & - & - & - & - & - & - & - & 87.5 & - & - & - & 90.7 & 74.5 & 91.1 & 89.4 \\
& Ersatz & - & - & 59.7 & - & - & - & - & - & - & 86.2 & - & - & 73.2 & - & 87.7 \\
& Punkt & - & - & - & - & - & - & - & - & - & 86.5 & - & 90.1 & 73.5 & 85.4 & 88.6 \\
\cmidrule{2-17}
& \ourunsup{} & 75.8 & 60.4 & 66.2 & 76.6 & 73.1 & 93.7 & 79.5 & 88.7 & - & 88.5 & 69.8 & 89.2 & 78.5 & 91.7 & 91.3 \\
& \ourthresholdtuned{} & 77.8 & 65.1 & 66.4 & 76.1 & 74.2 & 93.3 & 83.1 & 89.6 & - & 90.8 & 75.6 & 90.9 & 85.8 & 92.6 & 90.3 \\
& \ourpuncttuned{} & \textbf{88.5} & \textbf{72.0} & \textbf{77.2} & \textbf{83.8} & \textbf{89.8} & \textbf{96.5} & \textbf{87.4} & \textbf{94.5} & - & \textbf{95.2} & \textbf{82.6} & \textbf{95.0} & \textbf{90.1} & \textbf{96.2} & \textbf{95.0} \\
\midrule
\multirow{8}{*}{Ersatz} & \spacysent{} & - & - & 89.4 & - & - & - & - & - & - & 84.1 & - & - & 89.9 & - & 89.8 \\
& PySBD & - & - & 47.9 & - & - & - & - & - & - & - & - & - & 95.5 & - & 74.2 \\
& \spacydp{} & - & - & - & - & - & - & - & - & - & - & - & - & 96.3 & - & 98.3 \\
& Ersatz & - & - & 92.9 & - & - & - & - & - & - & 96.8 & - & - & 95.6 & - & 97.6 \\
& Punkt & - & - & - & - & - & - & - & - & - & 96.8 & - & - & 95.5 & - & 97.8 \\
\cmidrule{2-17}
& \ourunsup{} & - & - & 87.8 & - & - & - & - & - & - & 93.7 & - & - & 95.7 & - & 96.8 \\
& \ourthresholdtuned{} & - & - & 88.9 & - & - & - & - & - & - & 94.1 & - & - & 96.0 & - & 96.9 \\
& \ourpuncttuned{} & - & - & \textbf{92.9} & - & - & - & - & - & - & \textbf{98.9} & - & - & \textbf{99.3} & - & \textbf{98.7} \\
\bottomrule
\end{tabularx}
}
\caption{Sentence segmentation test F1 scores on languages af-en.}
\label{table:all_comparisons_first}
\end{table*}

\begin{table*}[t!]
\centering
\small
\setlength\tabcolsep{4pt}
\resizebox{\textwidth}{!}{
\begin{tabularx}{\linewidth}{llXXXXXXXXXXXXXXX}
\toprule
& & \thead{eo} & \thead{es} & \thead{et} & \thead{eu} & \thead{fa} & \thead{fi} & \thead{fr} & \thead{fy} & \thead{ga} & \thead{gd} & \thead{gl} & \thead{gu} & \thead{ha} & \thead{he} & \thead{hi}\\
\midrule
\multirow{8}{*}{UD} & \spacysent{} & - & 89.3 & 87.1 & 92.5 & 99.7 & 87.3 & 95.3 & - & 85.2 & - & - & - & - & 94.4 & 95.2 \\
& PySBD & - & 46.2 & - & - & 98.9 & - & 61.9 & - & - & - & - & - & - & - & 99.7 \\
& \spacydp{} & - & 99.0 & - & - & - & 95.5 & 92.0 & - & - & - & - & - & - & - & - \\
& Ersatz & - & 97.5 & 93.1 & - & - & 92.9 & 97.3 & - & - & - & - & - & - & - & 99.5 \\
& Punkt & - & 98.6 & 93.7 & - & - & 92.8 & 97.2 & - & - & - & - & - & - & - & - \\
\cmidrule{2-17}
& \ourunsup{} & - & 96.7 & 93.0 & 97.4 & 97.0 & 92.8 & 96.7 & - & 85.7 & 71.8 & 97.8 & - & - & 95.5 & 96.5 \\
& \ourthresholdtuned{} & - & 97.1 & 93.2 & 97.6 & 98.0 & 93.0 & 97.1 & - & 91.3 & 72.0 & \textbf{98.9} & - & - & 96.3 & 97.1 \\
& \ourpuncttuned{} & - & \textbf{99.7} & \textbf{98.2} & \textbf{99.9} & \textbf{99.9} & \textbf{98.2} & \textbf{98.8} & - & \textbf{98.1} & \textbf{81.2} & 98.6 & - & - & \textbf{97.2} & \textbf{99.9} \\
\midrule
\multirow{8}{*}{OPUS100} & \spacysent{} & - & 78.4 & 84.6 & 79.4 & 51.9 & 91.4 & 84.6 & - & 54.2 & - & - & 3.6 & - & 91.7 & 54.0 \\
& PySBD & - & 68.0 & - & - & 46.1 & - & 81.4 & - & - & - & - & - & - & - & 23.1 \\
& \spacydp{} & - & 88.4 & - & - & - & 92.9 & 84.6 & - & - & - & - & - & - & - & - \\
& Ersatz & - & 90.0 & 87.3 & - & - & 92.9 & 86.4 & - & - & - & - & 20.9 & - & - & 58.5 \\
& Punkt & - & 90.2 & 87.8 & - & - & 93.5 & 86.0 & - & - & - & - & - & - & - & - \\
\cmidrule{2-17}
& \ourunsup{} & 91.6 & 90.8 & 84.0 & 85.7 & 61.2 & 91.5 & \textbf{87.9} & 45.1 & 79.1 & 84.6 & 89.4 & 70.9 & 84.1 & 90.8 & 66.7 \\
& \ourthresholdtuned{} & 91.2 & 92.1 & 88.6 & 87.0 & 61.2 & 93.1 & - & 61.8 & 78.6 & 84.9 & 89.8 & 71.0 & 89.5 & 90.1 & 66.1 \\
& \ourpuncttuned{} & \textbf{95.7} & \textbf{95.4} & \textbf{94.9} & \textbf{92.2} & \textbf{73.7} & \textbf{96.1} & - & \textbf{88.6} & \textbf{87.9} & \textbf{92.8} & \textbf{94.4} & \textbf{77.8} & \textbf{92.1} & \textbf{94.1} & \textbf{77.5} \\
\midrule
\multirow{8}{*}{Ersatz} & \spacysent{} & - & 85.0 & 84.1 & - & - & 94.7 & 90.8 & - & - & - & - & 3.7 & - & - & 89.9 \\
& PySBD & - & 84.6 & - & - & - & - & 96.1 & - & - & - & - & - & - & - & 87.8 \\
& \spacydp{} & - & 96.4 & - & - & - & 95.2 & 87.6 & - & - & - & - & - & - & - & - \\
& Ersatz & - & 96.7 & \textbf{98.1} & - & - & 95.9 & 96.3 & - & - & - & - & 94.4 & - & - & \textbf{96.9} \\
& Punkt & - & 96.6 & 97.6 & - & - & 95.7 & 96.1 & - & - & - & - & - & - & - & - \\
\cmidrule{2-17}
& \ourunsup{} & - & 98.8 & 96.2 & - & - & 97.5 & 97.4 & - & - & - & - & 90.5 & - & - & 94.4 \\
& \ourthresholdtuned{} & - & 97.8 & 95.8 & - & - & 97.3 & 96.8 & - & - & - & - & 89.6 & - & - & 94.7 \\
& \ourpuncttuned{} & - & \textbf{99.5} & 98.0 & - & - & \textbf{99.4} & \textbf{98.6} & - & - & - & - & \textbf{96.9} & - & - & 96.4 \\
\bottomrule
\end{tabularx}
}
\caption{Sentence segmentation test F1 scores on languages eo-hi.}
\end{table*}

\begin{table*}[t!]
\centering
\small
\setlength\tabcolsep{4pt}
\resizebox{\textwidth}{!}{
\begin{tabularx}{\linewidth}{llXXXXXXXXXXXXXXX}
\toprule
& & \thead{hu} & \thead{hy} & \thead{id} & \thead{ig} & \thead{is} & \thead{it} & \thead{ja} & \thead{jv} & \thead{ka} & \thead{kk} & \thead{km} & \thead{kn} & \thead{ko} & \thead{ku} & \thead{ky}\\
\midrule
\multirow{8}{*}{UD} & \spacysent{} & 90.2 & 0.3 & 94.3 & - & 94.0 & 93.2 & 96.3 & - & - & - & - & - & - & - & - \\
& PySBD & - & 92.7 & - & - & - & 74.6 & 97.9 & - & - & 95.6 & - & - & - & - & - \\
& \spacydp{} & - & - & - & - & - & \textbf{99.6} & \textbf{98.0} & - & - & - & - & - & \textbf{99.9} & - & - \\
& Ersatz & - & - & - & - & - & - & 93.4 & - & - & 95.6 & - & - & - & - & - \\
& Punkt & - & - & - & - & - & 95.4 & - & - & - & - & - & - & - & - & - \\
\cmidrule{2-17}
& \ourunsup{} & 96.1 & 96.3 & \textbf{98.2} & - & 86.9 & 94.3 & 93.9 & \textbf{97.3} & - & \textbf{97.6} & - & - & 99.3 & - & - \\
& \ourthresholdtuned{} & 96.4 & 96.3 & - & - & 89.7 & 94.3 & 95.8 & - & - & 83.0 & - & - & 99.4 & - & - \\
& \ourpuncttuned{} & \textbf{99.3} & \textbf{98.1} & - & - & \textbf{96.7} & 99.5 & 98.0 & - & - & 97.2 & - & - & \textbf{99.9} & - & - \\
\midrule
\multirow{8}{*}{OPUS100} & \spacysent{} & 91.1 & 1.8 & 87.8 & - & 93.6 & 85.5 & 43.6 & - & - & - & - & 9.3 & - & - & 7.8 \\
& PySBD & - & 58.8 & - & - & - & 74.7 & 42.9 & - & - & 35.6 & - & - & - & - & - \\
& \spacydp{} & - & - & - & - & - & 85.7 & 42.2 & - & - & - & - & - & 47.4 & - & - \\
& Ersatz & - & - & - & - & - & - & 28.3 & - & - & 37.8 & 0.1 & - & - & - & - \\
& Punkt & - & - & - & - & - & 88.0 & - & - & - & - & - & - & - & - & - \\
\cmidrule{2-17}
& \ourunsup{} & 92.2 & \textbf{86.3} & 89.8 & 79.1 & 94.4 & 85.9 & 44.9 & - & 91.9 & 74.4 & 72.9 & 66.0 & 57.5 & 79.6 & 85.3 \\
& \ourthresholdtuned{} & 92.7 & - & 90.4 & 82.9 & 94.8 & 89.3 & 80.5 & - & 91.7 & 76.0 & 72.0 & 61.3 & 71.8 & 67.0 & 85.2 \\
& \ourpuncttuned{} & \textbf{96.5} & - & \textbf{94.5} & \textbf{90.7} & \textbf{96.9} & \textbf{94.0} & \textbf{87.4} & - & \textbf{93.2} & \textbf{92.5} & \textbf{79.3} & \textbf{78.5} & \textbf{82.6} & \textbf{84.8} & \textbf{90.9} \\
\midrule
\multirow{8}{*}{Ersatz} & \spacysent{} & - & - & - & - & - & - & 84.7 & - & - & - & - & - & - & - & - \\
& PySBD & - & - & - & - & - & - & 87.7 & - & - & 64.7 & - & - & - & - & - \\
& \spacydp{} & - & - & - & - & - & - & 91.2 & - & - & - & - & - & - & - & - \\
& Ersatz & - & - & - & - & - & - & 85.9 & - & - & 99.6 & 31.7 & - & - & - & - \\
& Punkt & - & - & - & - & - & - & - & - & - & - & - & - & - & - & - \\
\cmidrule{2-17}
& \ourunsup{} & - & - & - & - & - & - & 81.5 & - & - & 96.4 & 72.1 & - & - & - & - \\
& \ourthresholdtuned{} & - & - & - & - & - & - & 82.6 & - & - & 95.8 & 91.5 & - & - & - & - \\
& \ourpuncttuned{} & - & - & - & - & - & - & \textbf{94.8} & - & - & \textbf{99.8} & \textbf{92.0} & - & - & - & - \\
\bottomrule
\end{tabularx}
}
\caption{Sentence segmentation test F1 scores on languages hu-ky.}
\end{table*}

\begin{table*}[t!]
\centering
\small
\setlength\tabcolsep{4pt}
\resizebox{\textwidth}{!}{
\begin{tabularx}{\linewidth}{llXXXXXXXXXXXXXXX}
\toprule
& & \thead{la} & \thead{lt} & \thead{lv} & \thead{mg} & \thead{mk} & \thead{ml} & \thead{mn} & \thead{mr} & \thead{ms} & \thead{mt} & \thead{my} & \thead{ne} & \thead{nl} & \thead{no} & \thead{pa}\\
\midrule
\multirow{8}{*}{UD} & \spacysent{} & 89.2 & 88.0 & 92.4 & - & - & - & - & 68.0 & - & - & - & - & 90.7 & 93.5 & - \\
& PySBD & - & - & - & - & - & - & - & 60.0 & - & - & - & - & 93.6 & - & - \\
& \spacydp{} & - & 92.5 & - & - & - & - & - & - & - & - & - & - & 95.1 & - & - \\
& Ersatz & - & 92.6 & 96.8 & - & - & - & - & - & - & - & - & - & - & - & - \\
& Punkt & - & - & - & - & - & - & - & - & - & - & - & - & 95.6 & 95.5 & - \\
\cmidrule{2-16}
& \ourunsup{} & 89.5 & 98.3 & 96.5 & - & - & - & - & 90.5 & - & 90.6 & - & - & 94.4 & 98.2 & - \\
& \ourthresholdtuned{} & 90.5 & 98.1 & 96.4 & - & - & - & - & 92.8 & - & 88.0 & - & - & 93.6 & 98.5 & - \\
& \ourpuncttuned{} & \textbf{97.3} & \textbf{99.5} & \textbf{99.1} & - & - & - & - & \textbf{97.9} & - & \textbf{94.4} & - & - & \textbf{97.2} & \textbf{99.5} & - \\
\midrule
\multirow{8}{*}{OPUS100} & \spacysent{} & - & 67.2 & 58.0 & - & 90.4 & 39.9 & - & 84.4 & - & - & - & 15.3 & 92.2 & 92.1 & - \\
& PySBD & - & - & - & - & - & - & - & 86.1 & - & - & 27.2 & - & 18.2 & - & - \\
& \spacydp{} & - & 77.8 & - & - & 81.9 & - & - & - & - & - & - & - & 93.0 & - & - \\
& Ersatz & - & 77.3 & 77.6 & - & - & - & - & - & - & - & - & - & - & - & - \\
& Punkt & - & - & - & - & - & - & - & - & - & - & - & - & \textbf{93.9} & 94.8 & - \\
\cmidrule{2-17}
& \ourunsup{} & - & 78.3 & 79.6 & 90.0 & 93.0 & 81.3 & \textbf{81.0} & 89.1 & 87.7 & 63.0 & 70.5 & 70.6 & 92.2 & 94.7 & 56.3 \\
& \ourthresholdtuned{} & - & 85.3 & 86.5 & 92.1 & 93.0 & 82.4 & - & 89.0 & 88.5 & 81.1 & 75.5 & 70.2 & - & 94.8 & 63.3 \\
& \ourpuncttuned{} & - & \textbf{90.7} & \textbf{91.9} & \textbf{95.5} & \textbf{96.0} & \textbf{87.3} & - & \textbf{93.7} & \textbf{94.2} & \textbf{89.0} & \textbf{82.8} & \textbf{76.1} & - & \textbf{96.4} & \textbf{78.4} \\
\midrule
\multirow{8}{*}{Ersatz} & \spacysent{} & - & 74.3 & 89.8 & - & - & - & - & - & - & - & - & - & - & - & - \\
& PySBD & - & - & - & - & - & - & - & - & - & - & - & - & - & - & - \\
& \spacydp{} & - & 77.6 & - & - & - & - & - & - & - & - & - & - & - & - & - \\
& Ersatz & - & 95.1 & 98.7 & - & - & - & - & - & - & - & - & - & - & - & - \\
& Punkt & - & - & - & - & - & - & - & - & - & - & - & - & - & - & - \\
\cmidrule{2-17}
& \ourunsup{} & - & 96.9 & 97.2 & - & - & - & - & - & - & - & - & - & - & - & - \\
& \ourthresholdtuned{} & - & 96.6 & 97.3 & - & - & - & - & - & - & - & - & - & - & - & - \\
& \ourpuncttuned{} & - & \textbf{99.2} & \textbf{99.4} & - & - & - & - & - & - & - & - & - & - & - & - \\
\bottomrule
\end{tabularx}
}
\caption{Sentence segmentation test F1 scores on languages la-pa.}
\end{table*}

\begin{table*}[t!]
\centering
\small
\setlength\tabcolsep{4pt}
\resizebox{\textwidth}{!}{
\begin{tabularx}{\linewidth}{llXXXXXXXXXXXXXXX}
\toprule
& & \thead{pl} & \thead{ps} & \thead{pt} & \thead{ro} & \thead{ru} & \thead{si} & \thead{sk} & \thead{sl} & \thead{sq} & \thead{sr} & \thead{sv} & \thead{ta} & \thead{te} & \thead{tg} & \thead{th}\\
\midrule
\multirow{8}{*}{UD} & \spacysent{} & 93.0 & - & 86.2 & 98.6 & 76.6 & - & 85.3 & 92.7 & \textbf{100.0} & 74.4 & 90.0 & 92.3 & - & - & - \\
& PySBD & 85.0 & - & - & - & 67.7 & - & 86.6 & - & - & - & - & - & - & - & - \\
& \spacydp{} & 98.5 & - & \textbf{98.4} & 94.1 & 80.3 & - & - & - & - & - & 88.0 & - & - & - & - \\
& Ersatz & 97.5 & - & - & 98.3 & 78.3 & - & - & - & - & - & - & 91.0 & - & - & - \\
& Punkt & 97.4 & - & 92.0 & - & 78.2 & - & - & - & - & - & 94.2 & - & - & - & - \\
\cmidrule{2-17}
& \ourunsup{} & 94.9 & - & 96.1 & 82.5 & 86.1 & - & 96.2 & 96.0 & \textbf{100.0} & 97.9 & 95.1 & 97.2 & - & - & \textbf{69.5} \\
& \ourthresholdtuned{} & 95.8 & - & 95.7 & 94.0 & 87.7 & - & 96.0 & 96.5 & - & 98.2 & 95.3 & 97.9 & - & - & - \\
& \ourpuncttuned{} & \textbf{99.4} & - & 98.3 & \textbf{99.4} & \textbf{93.4} & - & \textbf{98.1} & \textbf{99.1} & - & \textbf{99.8} & \textbf{96.9} & \textbf{98.8} & - & - & - \\
\midrule
\multirow{8}{*}{OPUS100} & \spacysent{} & 89.4 & - & 90.1 & 90.7 & 72.4 & 75.8 & 88.1 & 89.7 & 87.6 & 91.6 & 90.8 & 40.8 & 62.5 & - & - \\
& PySBD & 17.6 & - & - & - & 65.9 & - & 29.5 & - & - & - & - & - & - & - & - \\
& \spacydp{} & 92.9 & - & 90.4 & 92.2 & 74.2 & - & - & - & - & - & 91.5 & - & - & - & - \\
& Ersatz & 92.2 & 1.6 & - & 92.8 & 68.7 & - & - & - & - & - & - & 45.2 & - & - & - \\
& Punkt & 92.8 & - & 92.4 & - & 75.9 & - & - & - & - & - & 92.9 & - & - & - & - \\
\cmidrule{2-17}
& \ourunsup{} & 92.4 & 64.2 & 90.7 & 89.5 & \textbf{82.1} & 80.4 & 90.7 & 92.0 & 89.7 & 94.3 & 91.3 & 66.1 & 78.7 & 81.5 & 68.8 \\
& \ourthresholdtuned{} & 92.8 & 71.6 & 92.1 & 90.0 & - & 80.8 & 93.1 & 93.3 & 90.8 & 94.8 & 93.1 & 66.5 & 78.7 & 83.8 & 71.5 \\
& \ourpuncttuned{} & \textbf{96.0} & \textbf{76.7} & \textbf{95.8} & \textbf{96.9} & - & \textbf{86.0} & \textbf{96.2} & \textbf{95.4} & \textbf{95.8} & \textbf{96.7} & \textbf{96.2} & \textbf{75.1} & \textbf{84.5} & \textbf{91.9} & \textbf{72.9} \\
\midrule
\multirow{8}{*}{Ersatz} & \spacysent{} & 77.6 & - & - & 89.6 & 88.3 & - & - & - & - & - & - & 88.9 & - & - & - \\
& PySBD & 46.1 & - & - & - & 55.4 & - & - & - & - & - & - & - & - & - & - \\
& \spacydp{} & 94.4 & - & - & 94.4 & 93.7 & - & - & - & - & - & - & - & - & - & - \\
& Ersatz & 95.1 & 93.7 & - & 95.9 & 94.3 & - & - & - & - & - & - & 95.6 & - & - & - \\
& Punkt & 94.3 & - & - & - & 93.7 & - & - & - & - & - & - & - & - & - & - \\
\cmidrule{2-17}
& \ourunsup{} & 94.8 & 85.0 & - & 97.8 & 97.6 & - & - & - & - & - & - & 94.6 & - & - & - \\
& \ourthresholdtuned{} & 92.8 & 91.6 & - & 97.1 & 97.7 & - & - & - & - & - & - & 95.0 & - & - & - \\
& \ourpuncttuned{} & \textbf{98.0} & \textbf{96.0} & - & \textbf{99.4} & \textbf{99.4} & - & - & - & - & - & - & \textbf{98.2} & - & - & - \\
\bottomrule
\end{tabularx}
}
\caption{Sentence segmentation test F1 scores on languages pl-th.}
\end{table*}

\begin{table*}[t!]
\centering
\small
\setlength\tabcolsep{4pt}
\resizebox{\textwidth}{!}{
\begin{tabularx}{\linewidth}{llXXXXXXXXXX}
\toprule
& & \thead{tr} & \thead{uk} & \thead{ur} & \thead{uz} & \thead{vi} & \thead{xh} & \thead{yi} & \thead{yo} & \thead{zh} & \thead{zu}\\
\midrule
\multirow{8}{*}{UD} & \spacysent{} & 94.9 & 90.7 & 99.1 & - & - & - & - & 77.5 & 96.7 & - \\
& PySBD & - & - & 99.1 & - & - & - & - & - & 98.9 & - \\
& \spacydp{} & - & 97.0 & - & - & - & - & - & - & 98.2 & - \\
& Ersatz & 97.6 & - & - & - & - & - & - & - & 88.8 & - \\
& Punkt & 96.3 & - & - & - & - & - & - & - & - & - \\
\cmidrule{2-12}
& \ourunsup{} & 94.3 & 92.6 & 92.4 & - & 67.8 & - & - & \textbf{84.5} & 98.1 & - \\
& \ourthresholdtuned{} & 94.3 & 93.1 & 96.1 & - & 91.4 & - & - & - & 98.2 & - \\
& \ourpuncttuned{} & \textbf{99.1} & \textbf{98.4} & \textbf{99.4} & - & \textbf{97.9} & - & - & - & \textbf{99.8} & - \\
\midrule
\multirow{8}{*}{OPUS100} & \spacysent{} & 91.6 & 86.5 & 40.0 & - & - & - & - & 14.1 & 64.1 & - \\
& PySBD & - & - & 30.8 & - & - & - & - & - & 69.8 & - \\
& \spacydp{} & - & 89.7 & - & - & - & - & - & - & 69.4 & - \\
& Ersatz & 92.7 & - & - & - & - & - & - & - & 54.7 & - \\
& Punkt & 93.6 & - & - & - & - & - & - & - & - & - \\
\cmidrule{2-12}
& \ourunsup{} & 93.4 & 89.1 & 53.8 & 78.2 & 90.7 & 78.7 & 74.7 & \textbf{76.9} & 81.0 & 73.2 \\
& \ourthresholdtuned{} & 93.6 & 89.8 & 53.0 & 80.4 & 90.9 & 81.9 & 75.7 & - & 77.8 & 83.9 \\
& \ourpuncttuned{} & \textbf{95.7} & \textbf{94.7} & \textbf{68.1} & \textbf{85.9} & \textbf{94.9} & \textbf{90.4} & \textbf{81.8} & - & \textbf{89.2} & \textbf{90.9} \\
\midrule
\multirow{8}{*}{Ersatz} & \spacysent{} & 85.5 & - & - & - & - & - & - & - & 90.6 & - \\
& PySBD & - & - & - & - & - & - & - & - & 92.7 & - \\
& \spacydp{} & - & - & - & - & - & - & - & - & 95.8 & - \\
& Ersatz & 96.3 & - & - & - & - & - & - & - & 87.6 & - \\
& Punkt & 92.9 & - & - & - & - & - & - & - & - & - \\
\cmidrule{2-12}
& \ourunsup{} & 93.4 & - & - & - & - & - & - & - & 93.5 & - \\
& \ourthresholdtuned{} & 93.4 & - & - & - & - & - & - & - & 93.7 & - \\
& \ourpuncttuned{} & \textbf{98.4} & - & - & - & - & - & - & - & \textbf{97.8} & - \\
\bottomrule
\end{tabularx}
}
\caption{Sentence segmentation test F1 scores on languages tr-zu.}
\label{table:all_comparisons_last}
\end{table*}

\newpage
\clearpage

\begin{table*}[!t]
\footnotesize
\def\arraystretch{0.88}
\centering
\scalebox{0.75}{%
\begin{tabularx}{\linewidth+2in}{ccc}
\toprule
\multicolumn{2}{c}{\multirow{6}{*}{Source Paragraph}} &  \makecell[Xt]{Higgins bat mich um einen Gefallen. Und ich fragte jemand anderen um einen Gefallen. Sie gravierten meinen Namen rein.\textcolor{red}{\textbf{ | }}Günstige und Luxus Hotels in Schillig:\textcolor{red}{\textbf{ | }}Stand: 07.11.201513:00:28\textcolor{red}{\textbf{ | }}Alle Wettbewerbsrunden sind öffentlich.\textcolor{red}{\textbf{ | }}Nicht jeder bekommt eine zweite Chance, Bruder.\textcolor{red}{\textbf{ | }}01/06/2011 :Auto Moto Rally: Runde Nsele 04 und 05 Juni 2011.\textcolor{red}{\textbf{ | }}Also, komm rein. Aber du musst ruhig sein, ja?\textcolor{red}{\textbf{ | }}32008 L 0057: Richtlinie 2008/57/EG des Europäischen Parlaments und des Rates vom 17. Juni 2008 über die Interoperabilität des Eisenbahnsystems in der Gemeinschaft (ABl. L 191 vom 18.7.2008, S. 1).“\textcolor{red}{\textbf{ | }}Gebiet: Tschechische Republik\textcolor{red}{\textbf{ | }}Was war daran auszusetzen?} \\
\cmidrule{3-3}
\multicolumn{2}{c}{\multirow{5}{*}{Target Paragraph}} & \makecell[Xt]{Higgins asked me a favor, I asked someone else a favor, they slapped my name on it.\textcolor{red}{\textbf{ | }}Accommodations in Schillig\textcolor{red}{\textbf{ | }}Stand: 07.11.201513:02:59\textcolor{red}{\textbf{ | }}All rounds of the competition are open to the public.\textcolor{red}{\textbf{ | }}Not everybody gets a second chance to do what's right, bro.\textcolor{red}{\textbf{ | }}01/06/2011 :Auto Moto Rally: Round of Nsele 04 and 05 June 2011.\textcolor{red}{\textbf{ | }}So, come on in, but keep it quiet, okay?\textcolor{red}{\textbf{ | }}Directive 2008/57/EC of the European Parliament and of the Council of 17 June 2008 on the interoperability of the rail system within the Community (OJ L 191, 18.7.2008, p. 1).’;\textcolor{red}{\textbf{ | }}Area: Czech Republic\textcolor{red}{\textbf{ | }}What was wrong with it?} \\
\midrule
\multirow{13}{*}{None} & \multirow{6}{*}{Segmentation} & \makecell[Xt]{Higgins bat mich um einen Gefallen. Und ich fragte jemand anderen um einen Gefallen. Sie gravierten meinen Namen rein. Günstige und Luxus Hotels in Schillig: Stand: 07.11.201513:00:28 Alle Wettbewerbsrunden sind öffentlich. Nicht jeder bekommt eine zweite Chance, Bruder. 01/06/2011 :Auto Moto Rally: Runde Nsele 04 und 05 Juni 2011. Also, komm rein. Aber du musst ruhig sein, ja? 32008 L 0057: Richtlinie 2008/57/EG des Europäischen Parlaments und des Rates vom 17. Juni 2008 über die Interoperabilität des Eisenbahnsystems in der Gemeinschaft (ABl. L 191 vom 18.7.2008, S. 1).“ Gebiet: Tschechische Republik Was war daran auszusetzen?} \\
\cmidrule{3-3}
& \multirow{5}{*}{Prediction (BLEU=47.1)} & \makecell[Xt]{Higgins asked me for a favor. And I asked someone else for a favor. They engraved my name in. Cheap and luxury hotels in Schillig: As of: 07.11.201513:00:28 All competition rounds are public. Not everyone gets a second chance, brother. 01/06/2011 :Auto Moto Rally: Round Nsele 04 and 05 June 2011. So, come in. But you have to be quiet, yes? 32008 L 0057: Directive 2008/57/EC of the European Parliament and of the Council of 17 June 2008 on the interoperability of the railway system in the Community (OJ L 191 of 18.7.2008, p. 1).} \\
\midrule
\multirow{13}{*}{Naïve} & \multirow{6}{*}{Segmentation} & \makecell[Xt]{Higgins bat mich um einen Gefallen. Und ich fragte \textcolor{red}{\textbf{ | }}jemand anderen um einen Gefallen. Sie gravierten meinen Namen rein. \textcolor{red}{\textbf{ | }}Günstige und Luxus Hotels in Schillig: Stand: 07.11.201513:00:28 Alle \textcolor{red}{\textbf{ | }}Wettbewerbsrunden sind öffentlich. Nicht jeder bekommt eine zweite Chance, Bruder. \textcolor{red}{\textbf{ | }}01/06/2011 :Auto Moto Rally: Runde Nsele 04 und 05 \textcolor{red}{\textbf{ | }}Juni 2011. Also, komm rein. Aber du musst ruhig sein, \textcolor{red}{\textbf{ | }}ja? 32008 L 0057: Richtlinie 2008/57/EG des Europäischen Parlaments \textcolor{red}{\textbf{ | }}und des Rates vom 17. Juni 2008 über die Interoperabilität \textcolor{red}{\textbf{ | }}des Eisenbahnsystems in der Gemeinschaft (ABl. L 191 vom \textcolor{red}{\textbf{ | }}18.7.2008, S. 1).“ Gebiet: Tschechische Republik Was war daran auszusetzen?} \\
\cmidrule{3-3}
& \multirow{5}{*}{Prediction (BLEU=46.6)} & \makecell[Xt]{Higgins asked me for a favor, and I asked\textcolor{red}{\textbf{ | }}Someone else for a favor, they engraved my name in.\textcolor{red}{\textbf{ | }}Cheap and Luxury Hotels in Schillig: As of: 07.11.201513:00:28 All\textcolor{red}{\textbf{ | }}Competition rounds are public. Not everyone gets a second chance, brother.\textcolor{red}{\textbf{ | }}01/06/2011 :Auto Moto Rally: Round Nsele 04 and 05\textcolor{red}{\textbf{ | }}So, come in, but you have to be quiet,\textcolor{red}{\textbf{ | }}yes? 32008 L 0057: Directive 2008/57/EC of the European Parliament\textcolor{red}{\textbf{ | }}and the Council of 17 June 2008 on interoperability\textcolor{red}{\textbf{ | }}of the rail system in the Community (OJ L 191,\textcolor{red}{\textbf{ | }}18.7.2008, p. 1).- Area: Czech Republic What had to be done about it?} \\
\midrule
\multirow{13}{*}{Ersatz} & \multirow{5}{*}{Segmentation} & \makecell[Xt]{Higgins bat mich um einen Gefallen. \textcolor{red}{\textbf{ | }}Und ich fragte jemand anderen um einen Gefallen. \textcolor{red}{\textbf{ | }}Sie gravierten meinen Namen rein. \textcolor{red}{\textbf{ | }}Günstige und Luxus Hotels in Schillig: Stand: 07.11.201513:00:28 Alle Wettbewerbsrunden sind öffentlich. \textcolor{red}{\textbf{ | }}Nicht jeder bekommt eine zweite Chance, Bruder. \textcolor{red}{\textbf{ | }}01/06/2011 :Auto Moto Rally: Runde Nsele 04 und 05 Juni 2011. \textcolor{red}{\textbf{ | }}Also, komm rein. \textcolor{red}{\textbf{ | }}Aber du musst ruhig sein, ja? \textcolor{red}{\textbf{ | }}32008 L 0057: Richtlinie 2008/57/EG des Europäischen Parlaments und des Rates vom 17. Juni 2008 über die Interoperabilität des Eisenbahnsystems in der Gemeinschaft (ABl. \textcolor{red}{\textbf{ | }}L 191 vom 18.7.2008, S. 1).“ Gebiet: Tschechische Republik Was war daran auszusetzen?} \\
\cmidrule{3-3}
& \multirow{6}{*}{Prediction (BLEU=52.1)} & \makecell[Xt]{Higgins asked me for a favor.\textcolor{red}{\textbf{ | }}And I asked someone else for a favor.\textcolor{red}{\textbf{ | }}They engraved my name in.\textcolor{red}{\textbf{ | }}Cheap and Luxury Hotels in Schillig: As of: 07.11.201513:00:28 All competition rounds are public.\textcolor{red}{\textbf{ | }}Not everyone gets a second chance, brother.\textcolor{red}{\textbf{ | }}01/06/2011 :Auto Moto Rally: Round Nsele 04 and 05 June 2011.\textcolor{red}{\textbf{ | }}So, come on in.\textcolor{red}{\textbf{ | }}But you have to be quiet, right?\textcolor{red}{\textbf{ | }}32008 L 0057: Directive 2008/57/EC of the European Parliament and of the Council of 17 June 2008 on the interoperability of the rail system in the Community (OJ L 347, 20.12.2008, p.\textcolor{red}{\textbf{ | }}OJ L 191, 18.7.2008, p. 1).} \\
\midrule
\multirow{13}{*}{\ourpuncttuned{}} & \multirow{6}{*}{Segmentation} & \makecell[Xt]{Higgins bat mich um einen Gefallen. \textcolor{red}{\textbf{ | }}Und ich fragte jemand anderen um einen Gefallen. \textcolor{red}{\textbf{ | }}Sie gravierten meinen Namen rein. \textcolor{red}{\textbf{ | }}Günstige und Luxus Hotels in Schillig: \textcolor{red}{\textbf{ | }}Stand: 07.11.201513:00:28 \textcolor{red}{\textbf{ | }}Alle Wettbewerbsrunden sind öffentlich. \textcolor{red}{\textbf{ | }}Nicht jeder bekommt eine zweite Chance, Bruder. \textcolor{red}{\textbf{ | }}01/06/2011 :Auto Moto Rally: Runde Nsele 04 und 05 Juni 2011. \textcolor{red}{\textbf{ | }}Also, komm rein. \textcolor{red}{\textbf{ | }}Aber du musst ruhig sein, ja? \textcolor{red}{\textbf{ | }}32008 L 0057: Richtlinie 2008/57/EG des Europäischen Parlaments und des Rates vom 17. Juni 2008 über die Interoperabilität des Eisenbahnsystems in der Gemeinschaft (ABl. L 191 vom 18.7.2008, S. 1).“ \textcolor{red}{\textbf{ | }}Gebiet: Tschechische Republik \textcolor{red}{\textbf{ | }}Was war daran auszusetzen?} \\
\cmidrule{3-3}
& \multirow{6}{*}{Prediction (BLEU=59.8)} & \makecell[Xt]{Higgins asked me for a favor.\textcolor{red}{\textbf{ | }}And I asked someone else for a favor.\textcolor{red}{\textbf{ | }}They engraved my name in.\textcolor{red}{\textbf{ | }}Cheap and Luxury Hotels in Schillig:\textcolor{red}{\textbf{ | }}Situation as at: 07.11.201513:00:28\textcolor{red}{\textbf{ | }}All competitions are open to the public.\textcolor{red}{\textbf{ | }}Not everyone gets a second chance, brother.\textcolor{red}{\textbf{ | }}01/06/2011 :Auto Moto Rally: Round Nsele 04 and 05 June 2011.\textcolor{red}{\textbf{ | }}So, come on in.\textcolor{red}{\textbf{ | }}But you have to be quiet, right?\textcolor{red}{\textbf{ | }}Directive 2008/57/EC of the European Parliament and of the Council of 17 June 2008 on the interoperability of the rail system in the Community (OJ L 191, 18.7.2008, p. 1).\textcolor{red}{\textbf{ | }}Area: Czech Republic\textcolor{red}{\textbf{ | }}What was wrong with that?} \\
\bottomrule
\end{tabularx}
}
\caption{Example paragraph of the German-English OPUS100 data with segmentations and translations following different strategies. Data in OPUS100 is shuffled, so the paragraph is not coherent; this has been shown by \citet{wicks-post-2021-unified} to have little impact on segmentation performance. The pipe ('\textcolor{red}{\textbf{|}}') indicates sentence boundaries.}
\label{table:mt_examples}
\end{table*}

\end{document}